\documentclass{article}

\usepackage[final]{neurips_2024}

\usepackage[utf8]{inputenc} 
\usepackage[T1]{fontenc}    
\usepackage{hyperref}       
\usepackage
{url}            
\usepackage{booktabs}       
\usepackage{amsfonts}       
\usepackage{nicefrac}       
\usepackage{microtype}      
\usepackage{xcolor}         

\usepackage{amsmath,amsfonts,bm}

\def\eqref#1{equation~\ref{#1}}

\def\1{\bm{1}}

\DeclareMathAlphabet{\mathsfit}{\encodingdefault}{\sfdefault}{m}{sl}
\SetMathAlphabet{\mathsfit}{bold}{\encodingdefault}{\sfdefault}{bx}{n}

\author{%
  Harshavardhan Kamarthi\\
  College of Computing\\
  Georgia Institute of Technology\\
  \texttt{harsha.pk@gatech.edu} \\
  \And
    B. Aditya Prakash\\
  College of Computing\\
  Georgia Institute of Technology\\
  \texttt{badityap@cc.gatech.edu} \\
}

\usepackage[english]{babel}
\usepackage[utf8]{inputenc} 
\usepackage[T1]{fontenc}    
\usepackage{hyperref}       
\usepackage{url}            
\usepackage{booktabs}       
\usepackage{amsfonts}       
\usepackage{nicefrac}       
\usepackage{microtype}      
\usepackage{xcolor}         

\usepackage{amsmath,wasysym}
\usepackage[ruled,linesnumbered,vlined]{algorithm2e}
\usepackage{graphicx}

\usepackage{subcaption}
\usepackage{multirow}
\usepackage[normalem]{ulem}
\useunder{\uline}{\ul}{}
\usepackage{textcomp}
\usepackage{footnote}
\usepackage{makecell}
\usepackage{xspace}
\usepackage{enumitem}
\usepackage{mdwlist}
\usepackage{wrapfig,lipsum,booktabs}
\usepackage{amsthm}

\iftrue
    \newcommand{\hk}[1]{\textcolor{orange}{[Harsha: #1 ]}}
    \newcommand{\aditya}[1]{\textcolor{red}{[Aditya: #1 ]}}
\else
    \newcommand{\hk}[1]{}
    \newcommand{\aditya}[1]{}
\fi

\newcommand{\model}{\textsc{LPTM}\xspace}
\newcommand{\taskpre}{{\mathcal{T}_{\text{pre}}}}
\def\datapre{{\mathcal{D}_{\text{pre}}}}
\newcommand{\randmask}[0]{\textsc{RandMask}\xspace}
\newcommand{\lastmask}[0]{\textsc{LastMask}\xspace}

\title{Large Pre-trained time series models for cross-domain Time series analysis tasks}

\begin{document}

\maketitle

\begin{abstract}
  Large pre-trained models have been vital in recent advancements in domains like language and vision, making model training for individual downstream tasks more efficient and provide superior performance.
However, tackling time-series analysis tasks usually involves
designing and training a separate model from scratch leveraging training data and domain expertise specific to the task.
We tackle a significant challenge for pre-training a foundational time-series model from multi-domain  time-series datasets:
extracting
semantically useful tokenized inputs to the model
across heterogenous time-series from different domains.
We propose Large Pre-trained Time-series Models (\model) that introduces a novel method of \textit{adaptive segmentation}
that automatically identifies optimal dataset-specific
segmentation strategy during pre-training.
This enables
\model to perform similar to or better than domain-specific state-of-art model
when fine-tuned to different downstream time-series analysis tasks and under zero-shot settings.
\model achieves superior forecasting and time-series classification results
taking up to 40\% less data and 50\% less training time
compared to state-of-art baselines.
Code: \url{www.github.com/AdityaLab/Samay}

\end{abstract}

\section{Introduction}

Time-series analysis tasks are important well-studied  problems  such as forecasting~\citep{hyndman2018forecasting} and classification~\citep{chowdhury2022tarnet} with applications in
wide-ranging domains such as retail, meteorology, economics, and health.
Recent works~\citep{chen2021autoformer, wang2022micn, zeng2023transformers} have shown the efficacy of purely data-driven deep learning models.
However, most state-of-art neural models are known to be data-hungry and require substantial training
data.
Motivated by language and vision foundational models~\cite{bommasani2021opportunities}, recent
body of works build
pre-trained time-series models~\cite{das2023decoder,ansari2024chronos,jin2023time,rasul2023lag}.
These models are trained on diverse datasets from different domains during pre-training.
They require less training resources and data and produce superior performance.
These models can also be deployed without any training, in a zero-shot or few-shot setting.

These foundational models, however, require large amounts of
data for pre-training, which is still a challenge
for time-series datasets.
Moreover, they do not provide consistent performance
improvement across all the domains.
We identify an important challenge to building a unified pre-trained model for time-series that is pre-trained on and deployed to multiple domains: representation
of diverse time-series input into these models.

Most neural sequential models input time-series values for each time-step separately.
However, unlike text data, each individual time stamp may not provide enough semantic meaning about local temporal patterns of the time series.
To tackle this, \cite{nie2022time} proposed to segment the time series and input each segment as individual tokens to their transformer-based model.
This simple segmentation method of \textit{tokenizing} time-series has been used by recent pre-trained models ~\cite{woo2024unified,das2023decoder}
to provide superior performance across multiple applications.
However, segmenting input time-series uniformly with fixed-length segments, while simple, can be a very inflexible
tokenizing method, especially
when dealing with datasets from
multiple domains with different set of underlying
generative dynamics, sampling rate, noise, etc.
For example, among two datasets, a dataset with a lower sampling rate (such as  GDP time-series) may require longer segments than those with higher sampling rates to capture similar patterns in the model (such as heart sensors collecting data in milliseconds).
However, note that the dynamics of the
same time-series may vary with time~\cite{liu2024timeood}.
For, time intervals that are smoother with less complex dynamics, using longer segment sizes may suffice but intervals where time-series are complex and have multiple temporal patterns may require finer-grained segmentation.
For example, for seasonal epidemics, time-series is smoother during the off-season and has more complex dynamics during outbreaks and times of higher incidence.

We tackle this important problem of representing diverse time-series datasets when training a
pre-trained foundational model for time-series
and propose \textbf{Large Pre-trained Time-series Models} (\model), a novel method for generating pre-trained models for time-series data across multiple domains.
\model uses a simple transformer-based architecture
and leverages masking-based self-supervised pre-training to train on multiple datasets
from different domains.
Our main contribution focuses on how we input time-series segments as tokens to the transformer.
To overcome the challenges associated with segmentation on diverse datasets discussed above, we propose a novel \textit{adaptive segmentation module} that segments the
time-series of each domain based on how well it performs on self-supervised pre-training.
The segmentation module uses a novel scoring mechanism during pre-training to identify an effective segmentation strategy for a domain.
\model can be fine-tuned or deployed in a zero-shot setting to various forecasting and classification tasks in domains such as epidemiology, energy, economics, behavioural datasets, etc.
\model provides performance on par with state-of-art models with lesser pre-training data, training data and fewer training steps. Our main contributions can be summarized as follows:

\noindent$\bullet$ \textbf{Multi-domain Pre-trained time-series model} We propose a framework for generating
foundational pre-trained models for time-series that are trained on multiple datasets across varied domains.
\model solves the tokenization problem for cross-domain time-series data and proposes a novel adaptive segmentation module which is important to build pre-trained models for time-series similar to foundational models for text and vision.

\noindent$\bullet$ \textbf{Adaptive segmentation for cross-domain pre-training} To optimally extract semantically useful information from time-series of different domains with varied dynamics and sampling rates for pre-training, we propose a novel adaptive segmentation module that learns segmentation strategy for each domain based on losses from self-supervised learning tasks.

\noindent$\bullet$ \textbf{State-of-art and efficient performance in diverse downstream time-series tasks} We evaluate \model on downstream forecasting and classification tasks from multiple domains and observe that \model consistently provides performance similar to or better than previous state-of-art models
usually under zero-shot evaluation as well as when fine-tuned with lesser training data and compute time. Overall, we also observe that \model typically requires less than 80\% of training data used by state-of-art baselines to provide similar or better performance.

\section{Problem Setup}
\paragraph{Time-series analysis tasks}
Our pre-trained model can be used for many time-series tasks including forecasting and classification
from multiple benchmarks and domains.
For a given downstream task let $\mathcal{D}^T$ be the time-series dataset consisting of time series $\mathbf{y}^{1\dots T}$.
A time-series analysis task's goal is to predict important properties of the time-series. For example, the forecasting task involves predicting the future values $\mathbf{y}^{T+1\dots T+K}$ whereas classification involves predicting the class label of the input time-series based on labelled training data.

\paragraph{Self-supervised pre-training on multi-domain datasets}
The goal of our work is to learn useful knowledge and patterns from time-series
datasets from time-series from different domains.
Formally, we have access to time-series datasets from $K$ domains where the datasets of domain $k$ is
denoted as $\mathcal{D}'_k = \{\mathcal{D}'_{k,i}\}_{i=1}^{N(k)}$ where $N(k)$ is the number of
datasets in domain $k$.
Examples of these domains include epidemiology, energy forecasting, macroeconomics, traffic prediction, etc.
The entire set of heterogenous multi-domain \textit{pre-train} dataset is denoted as $\datapre = \{\mathcal{D}'_1, \mathcal{D}'_2, \dots, \mathcal{D}'_K\}$.
In order to effectively pre-train \model on $\datapre$ we
formulate the problem as a set of self-supervised learning tasks
$\taskpre=\{\mathcal{T}_i\}_{i=1}^R$ on the set of pre-training datasets
$\datapre$.
During pre-training, we sample $(\mathcal{D}'_{k,i}, k)$, a dataset and its domain label from $\datapre$ and train the model
on each of the self-supervised learning tasks in $\taskpre$.
The tasks in $\taskpre$ are self-supervised and do not require additional labels
or other ground truth.
These tasks mask patches of the input data and train
the model to recover the original input.

Therefore, our problem can be formally stated as: \textit{Given a heterogeneous set of multi-domain datasets
	$\datapre$ and their domain labels, we train a model leveraging SSL tasks $\taskpre$ that learns
	important patterns and knowledge that can be leveraged on fine-tuning the model to any time-series analysis task on any novel dataset from any of the domains $d \in \{1, 2, \dots, K\}$.
}

\section{Methodology}

\begin{figure*}[h]
	\centering
	\includegraphics[width=.98\linewidth]{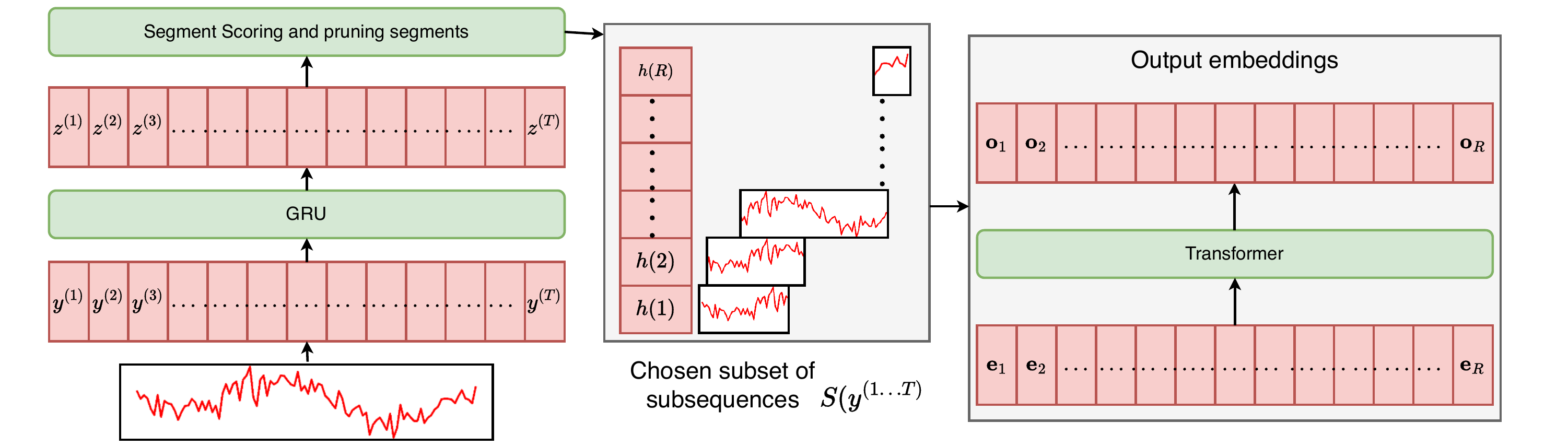}
	\caption{Overview of \model. The input time-series $y^{(1\dots T)}$ is first segmented based on a scoring function optimized using SSL loss. The segments are fed as individual tokens to the transformer encoder to  get output embeddings of time-series that are used for downstream tasks.}
	\label{fig:enter-label}
\end{figure*}

Similar to model pipelines used in NLP and vision, we first train a pre-trained model $M(\theta_{pre})$ on multiple pre-training datasets $\datapre$. Most parameters of the pre-trained model $\theta_{pre}$ are trained over all the datasets and tasks.
However, we use a separate segmentation module for each dataset domain to capture varied sizes of segments that differ across datasets.
These segments are used as tokens for a transformer model that shares the parameters across all the tasks. For each pre-trained and downstream task, we append a final linear layer on the output embeddings of the transformer to generate the final prediction.
Note that during fine-tuning on downstream tasks we update the parameters of all the modules of \model.

\subsection{Adaptive Segmentation module}

Previous works input each time-step of a time-series as individual tokens or fixed-sized segments.
Unlike text, individual time-steps do not typically provide any semantic meaning about the temporal patterns of the time-series.
Therefore, ~\cite{nie2022time} proposed to segment the input time-series  into uniform length segments and use each of the segments as tokens to the transformer model.
Segments of fixed size are also too inflexible to capture semantics of sequences that show varied behaviour across time and across datasets.
Different pre-trained datasets may have varied temporal scales, periodicity and other temporal dynamics that cannot be encompassed by a single uniform segmentation strategy.
For example, epidemic time-series are usually observed weekly and may have characteristic properties like seasonality, peaks and sudden outbreaks that should be captured by segmentation.
Economic time-series, in contrast, are typically captured  every quarter and are more monotone
with sudden anomalies and changes in data distribution.
Moreover, using a uniform segmentation may not be ideal for time series with multi-scale trends
with some time-stamps having denser temporal information requiring finer-graned segmentation than others.
Therefore, our goal is to identify an independent segmentation strategy for each domain of time-series dataset.

\paragraph{Overview}
\model automatically identifies an effective segmentation strategy for each dataset domain during pre-training.
The overarching goal of the segmentation module involves identifying segments that lead to lower SSL loss.
The segmentation module first scores all possible segments of the input time-series and the segments
with the highest scores are then chosen as the output segments used to tokenise the time-series.
The scoring function is trained such that the score of the segments will reflect how likely the chosen segments
will lower the SSL loss.

\paragraph{Details}
For a given input time-series $\mathbf{y}^{(1\dots t)}$, we pass it through
a GRU to get hidden embeddings $\{\mathbf{z}^{(i)}\}_{i=1}^t$ that models the temporal patterns of the input:
\begin{equation}
	\{\mathbf{z}^{(i)}\}_{i=1}^t = \text{GRU}_1(\{y^{(i)}\}_{i=1}^t).
	\label{eqn:gru}
\end{equation}
We then introduce a \textit{segment score function} $s$ that provides a scalar score for any subsequence of the input time-series:
\begin{equation}
	s(i,j) = \mathbf{v}^T\tanh \left(\mathbf{W}_1\mathbf{z}_{i} + \mathbf{W}_2\mathbf{z}_{j} + \mathbf{b} \right).
\end{equation}
Intuitively, the score $s(i,j)$ for a subsequence from time-stamp $i$
to $j$ denotes how good the given segment is for the dataset when optimizing for the SSL loss.

In next step, we sample subset $S(y^{(1\dots t)})$ of subsequences over the time-series that
a) covers the entire input time-series, b) has a high score function value.
While retrieving the optimal $S(y^{(1\dots t)})$ is an interesting combinatorial optimization problem,
we generate $S(y^{(1\dots t)})$ using a simple process as follows: for each $i \in \{1,2,\dots, t-1\}$, we denote $h(i) = \arg\max_{j \in \{i+1\dots, t-1\}} s(i, j)$
as the best segment starting from time-step $i$.
Then we generate the set of segments $\hat{S}(y^{(1\dots t)}) = \{(i, h(i))\}_{i=1}^{t-1}$.
To reduce the number of segments, we iteratively remove the lowest-scoring segments until we cannot remove any more segments without having time-steps not covered by any segments in the set.
The final set of segments after pruning is denoted as $S(y^{(1\dots t)})$.
The segmentation procedure is summarized in Alg. \ref{alg:segment}.

\newlength{\commentWidth}
\setlength{\commentWidth}{7cm}
\begin{algorithm}[h]
	\SetKwInOut{Input}{Input}
	\SetKwInOut{Output}{Output}
	\SetKwProg{myproc}{Procedure}{}{}
	\SetKwFunction{GetScores}{GetScores}
	\SetKwFunction{ChooseSegments}{ChooseSegments}
	\SetKwFunction{Training}{Training}
	\Input{Time-series $\mathbf{y}^{(1\dots t)} = \{y^{(i)}\}_{i=1}^t$}
	\myproc{\GetScores{$\mathbf{y}^{(1\dots t)}$}}{
	$\{\mathbf{z}^{(i)}\}_{i=1}^t$ $\gets$ GRU($\{y^{(i)}\}_{i=1}^t$)
	\tcp*{Encode time-series with GRU}
	\For{$i \in \{1, 2, \dots, t-1\}$}{
		\For{$j \in \{i+1, \dots, t\}$}{
			$s(i, j) \gets \mathbf{v}^T\tanh\left( \mathbf{W}_1\mathbf{z}^{(i)} + \mathbf{W}_2\mathbf{z}^{(j)} + \mathbf{b} \right)$
			\tcp*[l]{Scores for all segments}
		}
	}
	\KwRet $\{s(i,j)\}_{i,j \in \{1, \dots, t\}}^{i<j}$\;
	}
	\myproc{\ChooseSegments{$\{s(i,j)\}_{i,j \in \{1, \dots, t\}}^{i<j}$}}{
	\For{$i \in \{1, 2, \dots, t-1\}$}{
		$h(i) \gets \arg\max_{j\in \{i+1, \dots, t\}} s(i, j)$\tcp*{Best segment starting at index $i$}
	}
	$\hat{S} \gets \{(i, h(i))\}_{i=1}^{t-1}$\;
	$i \gets \arg\min_{j: (j, h(j)) \in \hat{S}} h(j)$\tcp*{Select lowest scoring segment}
	\While{Segments in $\hat{S} - \{(i, h(i))\}$ doesn't cover all time-steps $\{1, 2, \dots, t\}$}{
		$\hat{S} \gets \hat{S} - \{(i, h(i))\}$\;
		$i \gets \arg\min_{j: (j, h(j)) \in \hat{S}} h(j)$\;
	}
	\KwRet $S \gets \hat{S}$\;
	}
	\caption{Adaptive Segmentation Module}
	\label{alg:segment}
\end{algorithm}

To generate the token embeddings $\hat{\mathbf{e}}(i,j)$ for each segment $(i, j)$, we pass the embeddings $\{\mathbf{z}^{(i)}, \mathbf{z}^{(i+1)}, \dots, \mathbf{z}^{(j)}\}$ through a self-attention layer used in transformers and aggregate the output embeddings.
\begin{equation}
	\hat{\mathbf{e}}(i,j) = \sum_{k=i}^{j}\text{Self-Atten}\{\mathbf{z}^{(i)}, \mathbf{z}^{(i+1)}, \dots, \mathbf{z}^{(j)}\}
\end{equation}
Additionally, we concatenate the following features to the token embedding of each segment token to provide information regarding the position and length of the segment:
(1) Positional encoding of the starting time-step of the segment $pos(i)$ defined as:
\begin{equation}
	\text{pos}(i) = \begin{cases}
		\sin(i/10^{5i/D})     & \text{if }i \text{ is even} \\
		\cos(i/10^{5(i-1)/D}) & \text{if }i \text{ is odd.}
	\end{cases}
\end{equation}
where $D$ is the dimensions of output embedding of self-attention over $\{\hat{\mathbf{e}}_i, \hat{\mathbf{e}}_{i+1}, \dots, \hat{\mathbf{e}}_R\}$.
(2) Positional encoding of the length of the segment $pos(j-i)$.
The final output of the segmentation module is a sequence $\{\mathbf{e}_i\}_{i=1}^R$ where $R$ number of segments. The segments are arranged based on the ascending order of the first time-stamp of each segment.
The token embeddings are fed into a stack of transformer layers similar to the encoder of the transformer~\cite{vaswani2017attention}.
The output of the transformer layers is a sequence of output embeddings denoted as $\{\mathbf{o}_i\}_{i=1}^R$.

\subsection{Self-supervised learning Tasks}
Pre-training on a wide range of heterogeneous datasets from multiple domains helps \model learn from useful patterns and latent knowledge across these domains that can be generalized to range downstream tasks on different domains.
We propose two general self-supervised learning tasks motivated by pre-trained language models to enable \model to learn from all pre-trained datasets.
We leverage a transformer model and use the segment token embeddings of the segmentation module.
The two pre-training SSL tasks are \textbf{Random Masking} (\randmask) and \textbf{Last token masking} (\lastmask). \randmask allows the model to extrapolate and interpolate masked segments of the input time-series.
\randmask  has also been explored for representation learning in previous works~\citep{zerveas2021transformer,nie2022time} but are applied on the same dataset as that used for training unlike our data and task-agnostic pre-training setup.
Formally, we mask each input segment token with a probability of $\gamma$ and decode the values of time-series of the masked segments from the output embeddings of the transformer. We use a simple GRU with a single hidden layer on the output embedding of the masked token to decode the values of the segment. We use mean-squared error as the loss $\mathcal{L}_{SSL}$.
\lastmask is similar to \randmask except we mask last $\gamma$ fraction of the segments.
This allows the model to forecast the future values of the time-series, an important task in many time-series domains.

\subsection{Training details}
\label{sec:trainaddn}

\paragraph{Instance normalization}
The values of the time-series of each dataset can vary widely based on the time-series domain.
Therefore, as part of pre-processing, we first normalize the time-series of each dataset of pre-train datasets independently.
Moreover, the data distribution and the magnitude of the time-series can vary across time.
We use reversible instance normalization (REVIN) layer~\cite{kim2021reversible}.
REVIN performs instance normalization
on the input time-series and reverses the normalization of the output values.
The normalization step is part of the neural model and gradients are calculated over the normalization and reverse normalization layers.

\paragraph{Training the score function}
We use the loss from the SSL tasks to train the score function of the segmentation module and GRU in Eqn. \ref{eqn:gru}. Since there is no direct gradient flow between the score function and the final predictions, due to the discrete nature of choosing the segments, we match the aggregated scores of all the chosen segments in $S(y^{(1\dots t)})$ to the negative logarithm of the total MSE loss of both SSL tasks:
\begin{equation}
	\mathcal{L}_{g} = \left( \sum_{(i,j) \in S(y^{(1\dots t)})} s(i, j) + \log (\mathcal{L}_{SSL} )  \right)^2
\end{equation}
where $\mathcal{L}_{SSL}$ is the total loss of both SSL tasks. We also backpropagate over  $\mathcal{L}_{g}$ once every 10 batches to stabilize training since changing the segmentation strategy for every batch leads to unstable and inefficient training.

\paragraph{Linear-probing and fine-tuning}
\citet{Kumar2022FineTuningCD} showed that fine-tuning all the parameters of the pre-trained model for a specific downstream task can perform worse than just fine-tuning only the last layer (linear probing), especially for downstream tasks that are out-of-distribution to pre-trained data.
To alleviate this, they suggest performing a two-stage fine-tuning process: we first perform linear probing followed by fine-tuning all the parameters.

\section{Related Works}

\paragraph{Neural models for time-series analysis}
DeepAR~\cite{salinas2020deepar} is a popular forecasting model that trains an auto-regressive recurrent network to predict the parameters of the forecast distributions.
Deep Markov models \cite{krishnan2017structured,rangapuram2018deep, li2021learning, gu2021efficiently}  model the transition and emission components with neural networks. Recent works have also shown the efficacy of transformer-based models on
general time-series forecasting~\cite{oreshkin2019n,zhou2021informer,chen2021autoformer,zhou2022fedformer,liu2021pyraformer}.
However,  these methods do not perform pre-training and are trained independently for each application domain.
therefore, they do not leverage cross-domain
datasets to generate generalized models that can be used for a wide range of
benchmarks and tasks.

\paragraph{Self-supervised learning for time-series}
Recent works have shown the efficacy of
self-supervised representation learning for time-series for various classification
and forecasting tasks in a wide range of applications such as
modeling behavioral datasets~\cite{merrill2022self,chowdhury2022tarnet},
power generation~\cite{zhang2019deep}, health care~\cite{zhang2022self}.
\citet{franceschi2019unsupervised} used triplet loss to discriminate segments of the same time-series from others.
These works use methods such as contrastive losses~\cite{eldele2021time,yue2022ts2vec} or other similarity metric-based techniques~\cite{tonekaboni2021unsupervised}.
However, all these methods apply SSL on the same dataset that is used for training and may not adapt well to multi-domain datasets.
There has been some recent works leveraging foundational models like LLMs for time-series forecasting across multiple applications.
One set of works directly uses LLMs without fine-tuning to perform time-series forecasting via careful prompting~\cite{gruver2024large,jin2023time,liu2024lstprompt}.
Other works fine-tune LLMs specifically for time-series forecasting~\cite{zhou2023one,rasul2023lag,ansari2024chronos}.
The time-series representation used by these models includes using individual time-steps as input~\cite{ansari2024chronos}, converting each digit of the time-series to character embeddings to be directly used by LLMs~\cite{gruver2024large,jin2023time} and uniform segmentation~\cite{das2023decoder,woo2024unified,zhou2023one}.
\model performs superior to these methods while being 10x to 100x smaller than the large LLMs used as backbones.

\section{Experiment Setup}
\paragraph{Datasets}
\label{sec:pretraindata}
We derive pre-train time-series datasets from multiple domains:
$\bullet$ \textbf{Epidemics:} We use a large number of epidemic time-series aggregated by Project Tycho~\citep{van2018project}.
from 1888 to 2021 for different diseases collected at state and city levels in the US. We remove time series with missing data and use time series for 11 diseases of very diverse epidemic dynamics such as seasonality, biology, geography, etc.: Hepatitis A, measles, mumps, pertussis, polio, rubella,
smallpox, diphtheria, influenza, typhoid and Cryptosporidiosis (Crypto.).
$\bullet$ \textbf{Electricity:} We use ETT electricity datasets (ETT1 and ETT2) collected from \citep{zhou2021informer} at 1 hour intervals over 2 years. We use the default 12/4/4 train/val/test split and use the train split for pre-training as well.
$\bullet$ \textbf{Traffic Datasets:} We use 2 datasets related to traffic speed prediction.
PEMS-Bays (PEM-B) and METR-LA \citep{li2017diffusion} are datasets of traffic speed at various spots
collected by the Los Angeles Metropolitan Transportation Authority and California Transportation Agencies over 4-5 months.
$\bullet$ \textbf{Demand Datasets:} We use bike and taxi demand datasets (NY-B, NY-T) from New York City collected from April to June 2016 sampled every 30 minutes. We all but the last 5 days of data for training and pre-training.
$\bullet$ \textbf{Stock forecasting}: We also collect the time-series of daily stock prices of Nasdaq and S\&P 500 index using Yahoo finance data \citep{yfinance72:online} from July 2014 to June 2019. We train and pre-train using the first 800 trading days and use the last 400 for testing.
$\bullet$ \textbf{M4 competition time-series}: We also used the 3003 time-series of M4 forecasting competition~\citep{makridakis2000m3} which contains time-series from multiple domains including demographics, finance, and macroeconomics.
$\bullet$ \textbf{Motion and behavioral sensor datasets}: We use the set of sensor datasets extracted from UEA archive~\citep{bagnall2018uea} and UCI Machine learning repository~\citep{asuncion2007uci} similar to \citep{chowdhury2022tarnet}.
Note that our publicly accessible pre-training dataset is significantly smaller than other pre-trained datasets used by past work~\cite{ansari2024chronos,das2023decoder} some of which use confidential data inaccessible to us. We also do not use any synthetic datasets like~\cite{das2023decoder,ansari2024chronos}.

\paragraph{Downstream tasks}

We test the pre-trained \model trained on datasets discussed above on multiple forecasting and time-series classification tasks.
We perform forecasting on the influenza incidence time series in US and Japan. Specifically, we use the aggregated and normalized counts of outpatients exhibiting
influenza-like symptoms released weekly by CDC\footnote{\url{https://gis.cdc.gov/grasp/fluview/fluportaldashboard.html}}.
For influenza in Japan, we use influenza-affected patient counts collected by NIID\footnote{\url{https://www.niid.go.jp/niid/en/idwr-e.html}}. We forecast up to 4 weeks ahead over the period of 2004 to 2019 flu seasons using a similar setup as Flusight competitions~\cite{reich_collaborative_2019}.

We also perform electricity forecasting on the ETT1 and ETT2 datasets  using the train/test split mentioned previously. The last 10\% of PEM-Bays dataset is used for traffic forecasting up to 1 hour ahead and the last 5 days of New York demand datasets for demand forecasting up to 120 minutes in the future.
We also perform forecasting on the Nasdaq dataset for up to 5 days ahead and M3 time-series for 1 month ahead.
We use 6 of the sensor datasets from \cite{asuncion2007uci} for time-series classification tasks. We use an 80-20 train-test split similar to \cite{chowdhury2022tarnet}.

\paragraph{Baselines}
We compared \model's performance in various time-series tasks against twenty two state-of-the-art general forecasting and domain-specific baselines.
First, we compare against recent pre-trained foundational time-series models: (1) LLM-Time~\cite{gruver2024large}, (2) TimesFM~\cite{das2023decoder}, (3) Lag-LLAMA~\cite{rasul2023lag}, (4) Chronos~\cite{ansari2024chronos} and (5) MOIRAI~\cite{woo2024unified}.
We skip models like Time-LLM~\cite{jin2023time}, MOMENT~\cite{goswami2024moment} and Autotimes~\cite{liu2024autotimes} which cannot perform zero-shot forecasting across domains and are outperformed by the aforementioned recent models when fine-tuned.
We compared with (6) Informer~\cite{zhou2021informer}, (7) Autoformer~\cite{chen2021autoformer},
(8) iTransformer~\cite{liu2023itransformer} and
(9) PatchTST~\cite{nie2022time}, four state-of-the-art transformer-based forecasting models.
We also compare against other recent model (10) MICN~\citep{wang2022micn}, (11) TiDE~\cite{das2023long}
(12) TFT~\cite{lim2021temporal} and (13) TimesNeT~\cite{wu2022timesnet}.
We also compare with (14) N-HITS~\cite{challu2023nhits} which uses multi-scale interpolation and (15) AutoARIMA~\cite{hyndman2008automatic} a ARIMA based model that does automatic hyperparameter search.
We also compare it against three other state-of-art self-supervised methods for time-series:
(16) TS2Vec~\citep{yue2022ts2vec},
(17) TS-TCC~\citep{eldele2021time} and
(18) SimMTM~\cite{dong2024simmtm} uses masking as pre-trained task for time-series classification.

Finally, we compared against the best models for individual tasks for each domain. For influenza forecasting, we compared against previous state-of-art models (19) EpiFNP~\cite{kamarthi2021doubt} and (20) ColaGNN~\cite{deng2020cola} respectively.
We also compare against (21) STEP~\cite{shao2022pre} a SOTA model for demand forecasting, traffic prediction, and stock prediction benchmarks among the baselines by automatically modelling sparse relations between multiple features of the time-series.
For classification tasks, we compare against (22) TARNet~\cite{chowdhury2022tarnet}.

\section{Results}

\begin{table*}[h]
	\centering
	\caption{Average \textit{zero-shot} forecast performance (measured as RMSE over 10 runs) of \model and \textit{pre-trained} baselines. The best model is in \textbf{bold}. }
	\label{tab:zeroshot}
	\scalebox{0.8}{
		\begin{tabular}{c|ccccccccc}
			Model     & Flu-US     & Flu-japan & ETT1          & ETT2          & PEM-B        & NY-B         & NY-T        & Nasdaq     & M4             \\ \hline
			LLM-Time  & 1.38       & 1411      & 0.57          & 0.54          & 4.3          & 4.5          & 13.53       & 0.29       & 1.189          \\
			TimesFM   & 1.35       & 1259      & 0.61          & 0.59          & 3.9          & 3.9          & 13.11       & 0.29       & 1.211          \\
			Lag-LLAMA & 1.52       & 1488      & 0.83          & 1.06          & 5.3          & 3.8          & 12.84       & 0.24       & 1.311          \\
			Chronos   & 1.29       & 1274      & 0.62          & 0.56          & 4.2          & 3.6          & 13.74       & 0.29       & 1.125          \\
			MOIRAI    & 1.39       & 1411      & 0.69          & 0.52          & 4.2          & 4.4          & 13.82       & 0.27       & 1.192          \\
			TS2Vec    & 1.94       & 1233.1    & 1.33          & 1.82          & 3.7          & 4.1          & 14.39       & 0.87       & 1.616          \\
			TS-TCC    & 2.17       & 1356.15   & 1.14          & 1.57          & 4.1          & 3.8          & 15.72       & 0.92       & 1.492          \\
			SimMTM    & 2.17       & 1356.15   & 1.14          & 1.57          & 4.1          & 3.8          & 15.72       & 0.92       & 1.492          \\ \hline
			\model    & {\bf 1.14} & {\bf 996} & \textbf{0.53} & \textbf{0.49} & \textbf{3.4} & \textbf{3.2} & {\bf 13.12} & {\bf 0.22} & \textbf{0.972} \\\hline
		\end{tabular}
	}
\end{table*}
\begin{table}[h]
	\centering
	\caption{Average forecast performance (measured as RMSE over 10 runs) of \model and baselines over different domains. The best model is in \textbf{bold} and the second best is {\ul underlined}. }
	\label{tab:forecast}
	\scalebox{0.8}{
		\begin{tabular}{c|ccccccccc|c}
			Model               & Flu-US        & Flu-Japan & ETT1          & ETT2          & PEM-B        & NY-B          & NY-T           & Nasdaq        & M4             & Rank          \\ \hline
			AutoARIMA           & 2.14          & 1344      & 0.73          & 0.64          & 4.1          & 4.13          & 16.43          & 0.62          & 1.89           & 25.06         \\
			Informer            & 1.62          & 1139      & 0.57          & 0.71          & 3.1          & 2.89          & 12.33          & 0.83          & 1.055          & 15.89         \\
			Autoformer          & 1.41          & 1227      & 0.72          & 0.82          & 2.7          & 2.73          & 12.71          & 0.19          & 0.887          & 13.67         \\
			PatchTST            & 0.96          & 1113      & 0.52          & 0.63          & \textbf{2.5} & 2.64          & 11.95          & 0.15          & {\ul 0.877}    & 7.5           \\
			N-HITS              & 1.42          & 1211      & 0.53          & 0.62          & 2.9          & 2.74          & 11.87          & 0.57          & 0.968          & 13.0          \\
			TiDE                & 1.21          & 1186      & \textbf{0.49} & 0.49          & 3.5          & 3.86          & 11.95          & 0.57          & 1.078          & 13.44         \\
			MICN                & 0.95          & 1145      & \textbf{0.49} & {\ul 0.57}    & 3.6          & 2.61          & 11.56          & 0.13          & 0.931          & 7.77          \\
			TimesNet            & 1.04          & 1194      & 0.56          & 0.62          & 3.9          & 2.83          & 11.82          & 0.19          & 1.055          & 12.11         \\
			TFT                 & 1.21          & 1876      & 0.52          & 0.51          & 4.6          & 2.95          & 12.55          & 0.24          & 1.18           & 16.11         \\
			iTransformer        & 1.14          & 1256      & 0.57          & 0.59          & 4.3          & 2.83          & 13.16          & 0.29          & 1.125          & 17.5          \\
			LLM-Time            & 1.21          & 1319      & 0.52          & 0.49          & 3.9          & 3.7           & 12.11          & 0.21          & 1.064          & 14.05         \\
			TimesFM             & 1.32          & 1214      & 0.58          & 0.49          & 3.7          & 2.8           & 12.19          & 0.22          & 1.07           & 13.44         \\
			Lag-LLAMA           & 1.46          & 1416      & 0.61          & 0.57          & 3.9          & 2.9           & 13.43          & 0.28          & 1.33           & 20.16         \\
			Chronos             & 1.21          & 1228      & 0.59          & 0.52          & 3.7          & 3.1           & 12.82          & 0.27          & 1.04           & 15.55         \\
			MOIRAI              & 1.31          & 1336      & 0.62          & 0.55          & 3.9          & 3.5           & 13.71          & 0.24          & 1.21           & 19.22         \\
			STEP                & 1.17          & 983       & 0.54          & 0.93          & 2.7          & {\ul 2.52}    & \textbf{10.37} & \textbf{0.11} & 1.331          & 10.33         \\
			EpiFNP              & \textbf{0.52} & 872       & 0.81          & 1.25          & 4.1          & 2.98          & 12.11          & 0.28          & 1.281          & 16.77         \\
			ColaGNN             & 1.65          & 694       & 0.72          & 1.19          & 3.9          & 3.19          & 14.97          & 0.25          & 1.185          & 19.22         \\
			TS2Vec              & 1.85          & 905.9     & 0.99          & 1.74          & 3.5          & 3.11          & 13.48          & 0.94          & 1.344          & 21.94         \\
			SimMTM              & 1.31          & 1289      & 0.61          & 0.55          & 3.4          & 3.1           & 12.79          & 0.28          & 1.284          & 17.94         \\
			TS-TCC              & 1.94          & 1134.6    & 0.75          & 1.29          & 3.3          & 2.97          & 15.55          & 0.76          & 1.274          & 21            \\ \hline
			\model              & {\ul 0.79}    & {\ul 704} & \textbf{0.49} & \textbf{0.46} & \textbf{2.5} & \textbf{2.37} & {\ul 11.84}    & {\ul 0.17}    & \textbf{0.872} & \textbf{2.55} \\
			\model-NoSegment    & 0.93          & 766       & 0.57          & 0.55          & 3.2          & 3.17          & 14.96          & 0.27          & 1.146          & 13.72         \\
			\model-NoPreTrain   & 0.96          & 827       & 0.46          & 0.57          & 3.7          & 2.66          & 12.43          & 0.25          & 1.271          & 11.66         \\
			\model-NoLinProb    & 0.92          & 885       & 0.43          & 0.53          & 3.1          & 2.49          & 12.17          & 0.19          & 1.032          & 6.55          \\
			\model-OnlyRandMask & 0.87          & 895       & 0.51          & 0.52          & 2.8          & 2.42          & 12.36          & 0.21          & 1.076          & 8.0           \\
			\model-OnlyLastMask & 0.79          & 773       & 0.44          & 0.48          & 2.7          & 2.31          & 12.04          & 0.19          & 0.916          & \textbf{3.77} \\ \hline
		\end{tabular}
	}
\end{table}
The code for implementation of \model and datasets are provided at anonymized link\footnote{\url{www.github.com/AdityaLab/Samay}} and hyperparameters are discussed in the Appendix.
\model consists of 10 layers for the transformer and overall has about 100M parameters which 2x to over 10x smaller than other pre-trained time-series models.
\paragraph{Zero-shot forecasting}
An important benefit of foundational models in language and vision is their ability to adapt to novel
tasks without any fine-tuning in a zero-shot setting~\cite{brown2020language}.
We evaluate the zero-shot performance of \model and other
pre-trained baselines.
Similar to~\cite{gruver2024large} we use the last 20\% of the datasets for zero-shot evaluation.
We do not fine-tune the models but only input the normalized time-series for new tasks directly during inference.
The results are summarized in Table~\ref{tab:zeroshot}.
\model outperforms all the baselines significantly.
Moreover, the baselines such as TS2Vec, TS-TCC and SimMTM which are not designed to handle datasets from multiple domains perform much worse than other pre-trained methods.
This shows the importance of adaptive segmentation-based tokenization of \model to better generalize across multiple domains.

\paragraph{Fine-tuned forecasting}
We also train the models with training data to evaluate fine-tuned performance.
The forecasting performance are shown in Table~\ref{tab:forecast}.
We evaluated \model against seventeen other forecasting baselines.
\model is either the first or a close second best-performing model in all the benchmarks.
\model generally outperforms
general time-series forecasting methods as well as
2recent pre-trained models in all benchmarks despite its much lower parameter count (2-10x lower).
Further, it is competitive or superior in performance
to domain-specific
methods designed specifically for the given domains (such as EpiFNP and STEP).
\model beats the previous state-of-art domain-specific baselines in five of the benchmarks and comes second in four more. Finally, \model improves upon the state-of-art on electricity forecasting, traffic forecasting, and M4 datasets.

\paragraph{Time-series classification}
Unlike other autoregressive foundational models designed for forecasting, \model can be used for classification due to its encoder-style architecture.
We add a single classification layer over pooled output embeddings $\{\mathbf{o}^{(i)}\}_{i=1}^R$ to predict the class logits.
We evaluate \model and baselines on the classification of 35 sensor and behavioral datasets from UCI classification repositiry\citep{asuncion2007uci}. We report the accuracy scores in Table~\ref{tab:classify}. We observe that \model
has highest mean rank and largest number of times it outperforms all baselines.

\begin{table*}[h]
	\centering
	\caption{Average classification performance (measured as accuracy score over 10 runs) of \model and baselines over different domains. The best model is in \textbf{bold} and the second best is {\ul underlined}. The best model is statistically significant over the baselines ($p\leq0.05$) when it beats the previous state-of-art.}
	\label{tab:classify}
	\scalebox{0.70}{
		\begin{tabular}{c|cccccccccc}
			                          & Informer & Autoformer & TimesNet & TARNet        & TS2Vec     & TS-TCC        & TST        & SimMTM     & CRT        & \model        \\ \hline
			BasicMotions              & 0.95     & 0.93       & 0.92     & \textbf{1.00} & 0.99       & \textbf{1.00} & 0.92       & 0.86       & 0.88       & \textbf{1.00} \\
			FaceDetection             & 0.51     & 0.49       & 0.59     & {\ul 0.63}    & 0.51       & 0.54          & 0.62       & 0.73       & 0.78       & \textbf{0.79} \\
			FingerMovements           & 0.58     & 0.54       & 0.58     & {\ul 0.62}    & 0.46       & 0.47          & 0.59       & 0.68       & 0.72       & \textbf{0.78} \\
			PEMS-SF                   & 0.67     & 0.71       & 0.84     & \textbf{0.94} & 0.75       & 0.73          & 0.93       & 0.86       & 0.89       & {\ul 0.93}    \\
			RacketSports              & 0.83     & 0.86       & 0.91     & \textbf{0.98} & 0.77       & 0.85          & 0.79       & 0.84       & 0.87       & {\ul 0.93}    \\
			EigenWorms                & 0.49     & 0.62       & 0.73     & {\ul 0.89}    & 0.84       & 0.77          & 0.72       & 0.82       & 0.79       & \textbf{0.94} \\
			ArticularyWordRecognition & 0.83     & 0.82       & 0.79     & {\ul 0.97}    & 0.89       & 0.97          & 0.92       & 0.92       & 0.88       & {\bf 0.98}    \\
			AtrialFibrillation        & 0.57     & 0.55       & 0.68     & {\bf 1.00}    & 0.44       & 0.37          & 0.72       & 0.85       & 0.89       & {\ul 0.93}    \\
			CharacterTrajectories     & 0.57     & 0.55       & 0.68     & { 0.97}       & 0.98       & 0.96          & {\bf 0.99} & 0.97       & 0.93       & {\ul 0.98}    \\
			Cricket                   & 0.94     & 0.87       & 0.88     & { \bf 1.00}   & 0.98       & 0.97          & 0.84       & 0.96       & 0.94       & {\ul 0.99}    \\
			DuckGeese                 & 0.54     & 0.44       & 0.56     & { \ul 0.75}   & 0.39       & 0.57          & 0.74       & 0.58       & 0.55       & {\bf 0.79}    \\
			Epilepsy                  & 0.58     & 0.57       & 0.61     & { \bf 1.00}   & 1.00       & 0.98          & 0.94       & 0.78       & 0.55       & {\ul 0.97}    \\
			ERing                     & 0.23     & 0.29       & 0.46     & { \ul 0.92}   & 0.89       & 0.78          & 0.86       & 0.64       & 0.75       & {\bf 0.97}    \\
			EthanolConcentration      & 0.14     & 0.27       & 0.34     & {  0.32}      & 0.45       & 0.37          & {\ul 0.46} & 0.34       & 0.21       & {\bf 0.53}    \\
			HandMovementDirection     & 0.14     & 0.27       & 0.34     & {  0.32}      & 0.45       & 0.37          & {\ul 0.46} & 0.34       & 0.21       & {\bf 0.53}    \\
			Handwriting               & 0.16     & 0.18       & 0.29     & {  0.24}      & 0.52       & 0.55          & { 0.37}    & {\bf 0.64} & 0.52       & { 0.51}       \\
			Heartbeat                 & 0.53     & 0.66       & 0.61     & { \bf 0.78}   & 0.71       & 0.69          & { 0.74}    & 0.74       & 0.62       & {\ul 0.74}    \\
			InsectWingbeat            & 0.13     & 0.16       & 0.14     & {  0.14}      & 0.18       & 0.22          & {\ul 0.69} & 0.62       & 0.18       & {\ul 0.72}    \\
			JapaneseVowels            & 0.87     & 0.96       & 0.94     & { \bf 0.99}   & 0.97       & 0.98          & {\bf 0.99} & 0.94       & {\bf 0.99} & {\ul 0.98}    \\
			Libras                    & 0.72     & 0.64       & 0.75     & {\bf 1.00}    & 0.85       & 0.92          & { 0.91}    & 0.82       & 0.88       & {\ul 0.95}    \\
			LSST                      & 0.36     & 0.44       & 0.32     & {\ul  0.97}   & 0.54       & 0.62          & { 0.65}    & 0.53       & 0.48       & {\bf 0.98}    \\
			MotorImagery              & 0.51     & 0.52       & 0.50     & {\bf  0.64}   & 0.62       & 0.61          & {\ul 0.62} & 0.46       & 0.58       & { 0.57}       \\
			NATOPS                    & 0.75     & 0.69       & 0.84     & {\ul  0.92}   & 0.91       & 0.92          & {\bf 0.94} & 0.85       & 0.88       & {\bf 0.94}    \\
			PenDigits                 & 0.84     & 0.86       & 0.81     & {\ul  0.97}   & {\bf 0.98} & 0.95          & {\ul 0.97} & 0.94       & 0.88       & { 0.92}       \\
			Phoneme                   & 0.11     & 0.13       & 0.15     & {  0.17}      & 0.28       & 0.226         & {\bf 0.29} & 0.26       & 0.18       & {\bf 0.32}    \\
			SelfRegulation            & 0.65     & 0.76       & 0.57     & {  0.81}      & 0.77       & 0.84          & { 0.89}    & {\bf 0.96} & 0.73       & {\ul 0.92}    \\
			SpokenArabicDigits        & 0.92     & 0.96       & 0.92     & {  0.98}      & 0.94       & 0.97          & {\ul 0.99} & 0.96       & 0.94       & {\bf 1.00}    \\
			StandWalkJump             & 0.21     & 0.10       & 0.34     & {  0.53}      & 0.28       & 0.31          & {\bf 0.61} & {\ul 0.52} & 0.11       & {\ul 0.58}    \\
			UWaveGesture              & 0.79     & 0.86       & 0.82     & {  0.87}      & 0.91       & 0.92          & {0.86}     & 0.74       & 0.82       & {\bf 0.94}    \\
			PAMAP2                    & 0.73     & 0.87       & 0.84     & { \bf 0.97}   & 0.93       & 0.94          & {\ul 0.96} & 0.93       & 0.88       & {\bf 0.97}    \\
			OpportunityGestures       & 0.73     & 0.66       & 0.68     & { \ul 0.83}   & 0.92       & 0.74          & { 0.74}    & 0.62       & 0.72       & {\bf 0.92}    \\
			OpportunityLocomotion     & 0.84     & 0.78       & 0.85     & { \bf 0.91}   & 0.75       & 0.82          & { 0.84}    & 0.87       & 0.89       & {\ul 0.89}    \\
			SelfRegulationSCP2        & 0.562    & 0.598      & 0.612    & 0.622         & 0.593      & 0.575         & 0.604      & 0.614      & 0.625      & 0.691         \\
			Occupancy                 & 0.774    & 0.739      & 0.814    & 0.833         & 0.876      & 0.865         & 0.881      & 0.826      & 0.814      & 0.836         \\
			MosquitoSound             & 0.439    & 0.493      & 0.551    & 0.632         & 0.649      & 0.662         & 0.691      & 0.703      & 0.624      & 0.715         \\
		\end{tabular}
	}
\end{table*}

\paragraph{Data efficiency}
A significant advantage of leveraging pre-trained models
is that we do not require large datasets for fine-tuning to a specific task.
We evaluate the efficacy of \model to train with a fraction of training data.
For each time-series analysis task, we fine-tune the model using only $k\%$ of training
data for different values of $k$. We use the first $k\%$ of the timestamps' values. We do not choose a random sample to prevent data mixing from the rejected portion of training data.
We also performed the similar experiment on the best baseline for each task
and compare data efficiency of baseline with \model.
\begin{figure*}[htbp]
	\centering
	\begin{subfigure}{0.24\textwidth}
		\includegraphics[width=\linewidth]{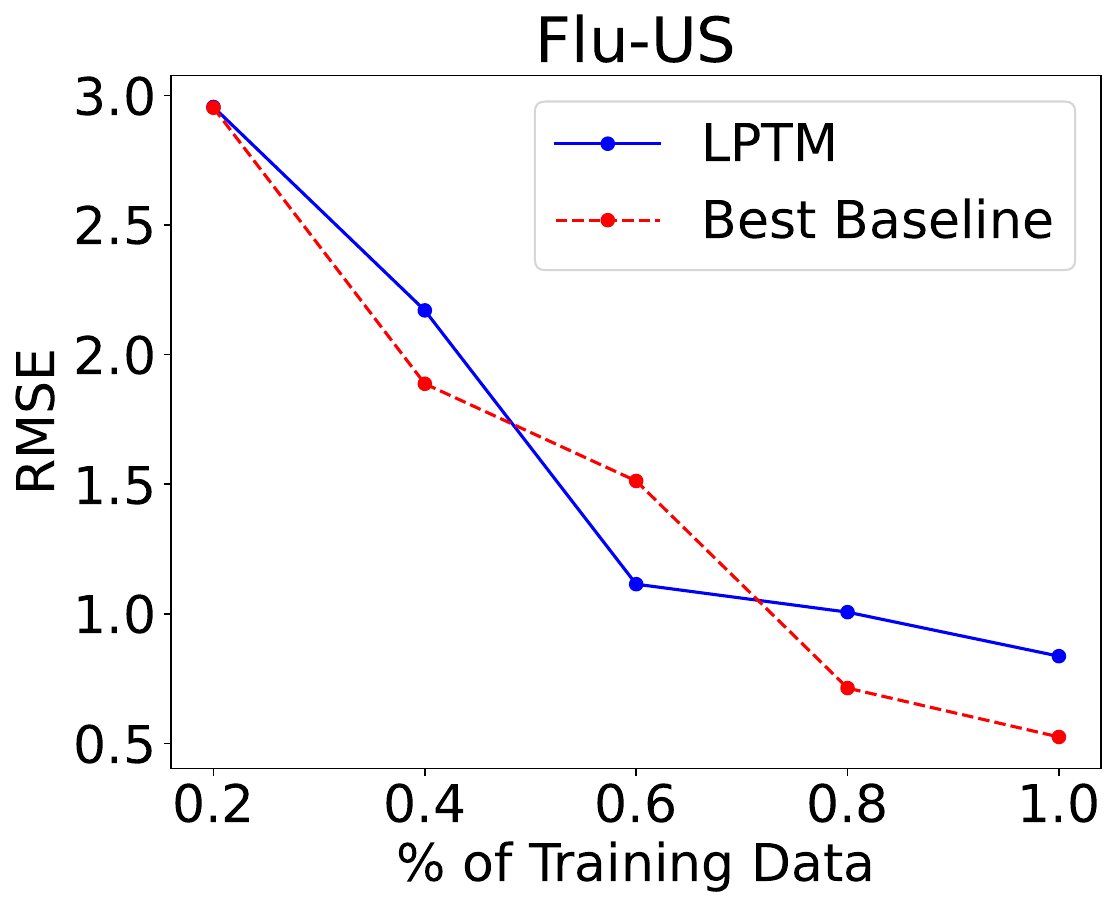}
	\end{subfigure}
	\hfill
	\begin{subfigure}{0.24\textwidth}
		\includegraphics[width=\linewidth]{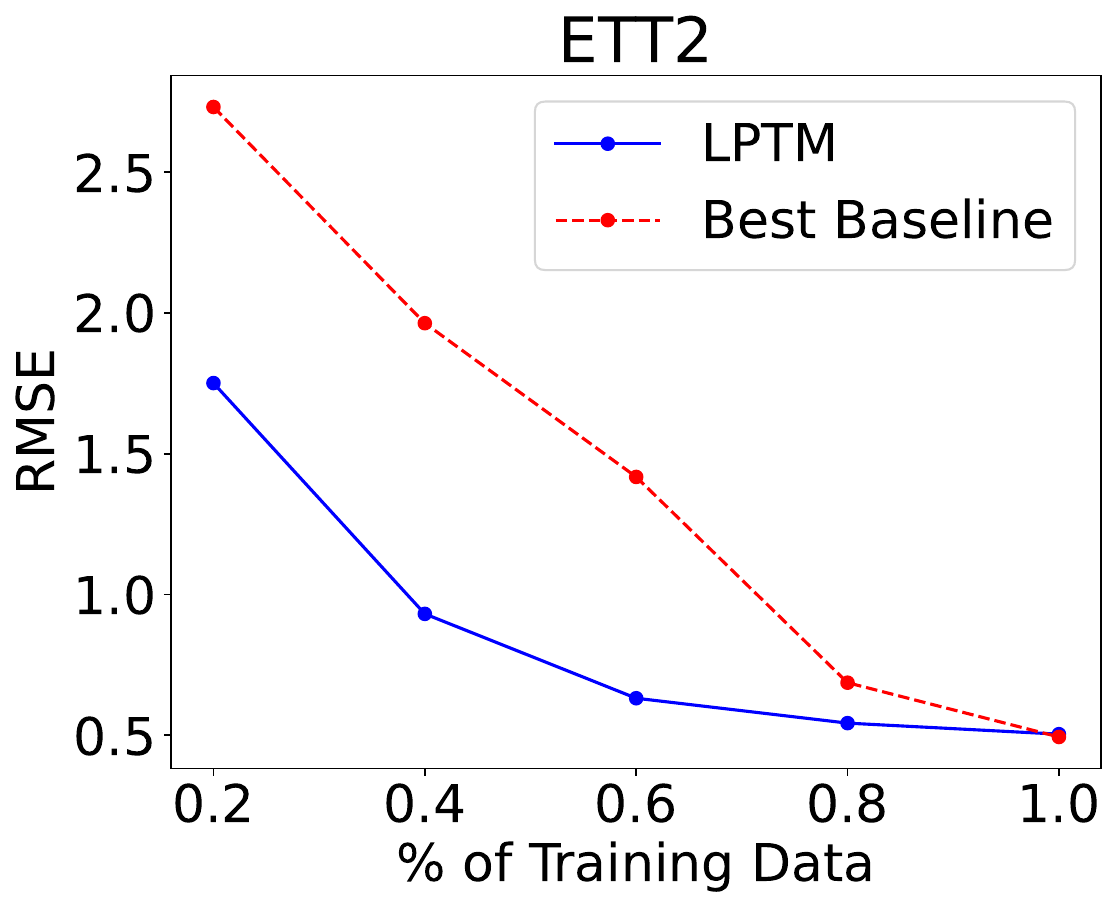}
	\end{subfigure}
	\hfill
	\begin{subfigure}{0.24\textwidth}
		\includegraphics[width=\linewidth]{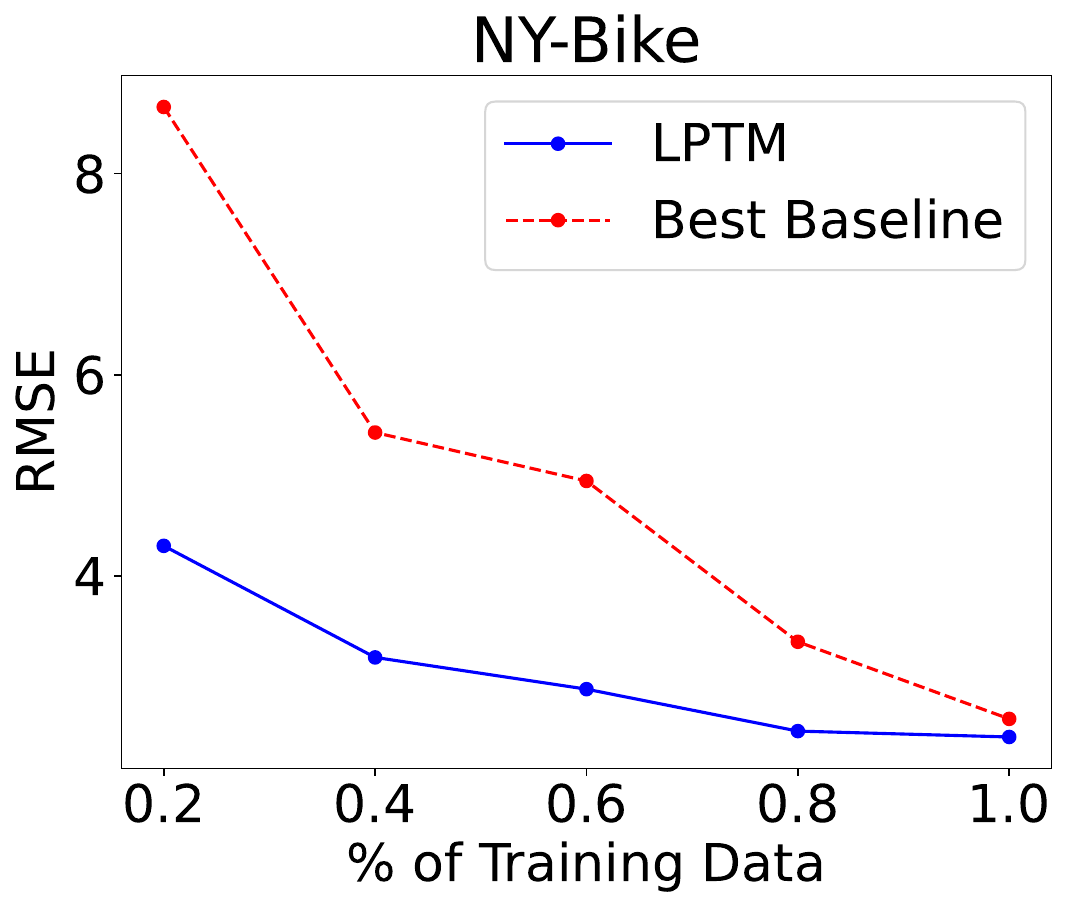}
	\end{subfigure}
	\hfill
	\begin{subfigure}{0.24\textwidth}
		\includegraphics[width=\linewidth]{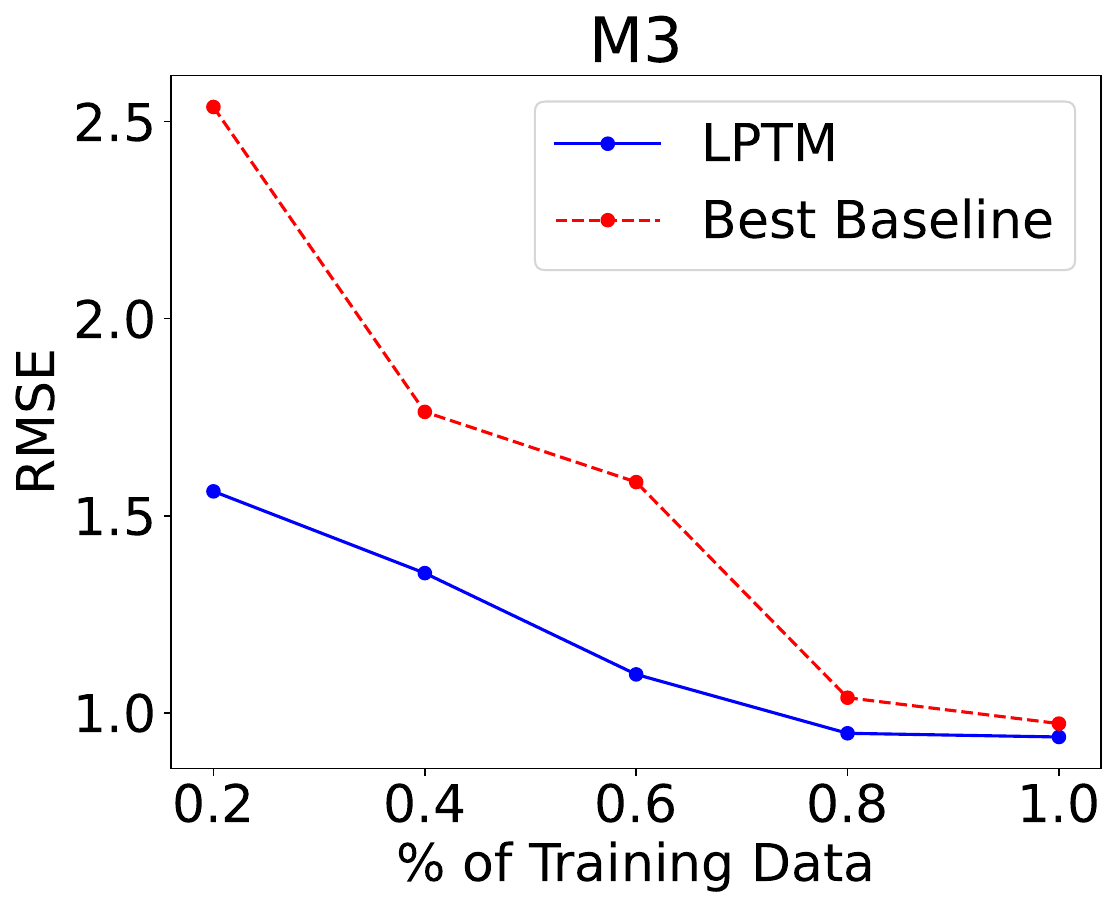}
	\end{subfigure}

	\caption{Performance of \model and best baseline with varying fractions of training data.  In most cases \model significantly outperforms baselines with lower amount of data.}
	\label{fig:dataeff}
\end{figure*}
The comparison plots are shown in Figure \ref{fig:dataeff}.
With lesser data, the performance of the baseline is much worse whereas \model typically requires much less data to provide similar performance to when we have access to the full dataset. This shows the importance of pre-training to quickly ramp up the performance of the model with much less data, a problem we encounter is many real-world settings such as when we need to deploy a forecasting model on novel applications such as a new pandemic with sparse data availability.

\paragraph{Training efficiency}
An important advantage of pre-trained models is that they require
much less training time and resources to fine-tune to a downstream task
compared to time required for pre-training or even training from scratch.
We compare the fine-tuning time of \model with baselines on benchmarks from different domains. We also measure the average time required by \model to reach the performance of the best baseline in cases where we eventually outperform them.
The training times are summarized in Appendix Table \ref{tab:traintime}.
We observe that the time taken by \model to reach the performance of best best-performing baseline (\model-TB) is significantly smaller than the time taken by any other baselines.
Even when \model doesn't outperform the best baseline, it typically converges much faster.

\paragraph{Ablation and Sensitivity Studies}
We study the impact of our adaptive segmentation as well as pre-training and linear probing via the ablation models \model-NoSegment, \model-NoPreTrain and
\model-NoLinProb.
We also investigate the individual impact of both SSL task via the ablation models \model-OnlyRandMask and \model-OnlyLastMask.
The performance of the ablation variants are also shown in Tables \ref{tab:forecast}.
We observe that the ablation variants' performances are significantly worse than the variants,  underperforming some of the baselines.
The worst performing variant is usually \model-NoSegment, showing the importance of deriving good time-series segments to improve representation learning of time-series for each dataset.
We also examined the sensitivity of hyperparameter $\gamma$ for SSL tasks and found the optimal value at 0.4 for LASTMASK and 0.2 for RANDMASK. The sensitivity analysis plots are in Appendix Fig. \ref{fig:gamma}.

\begin{figure}[h]
	\centering
	\begin{subfigure}[b]{0.32\textwidth}
		\centering
		\includegraphics[width=\textwidth]{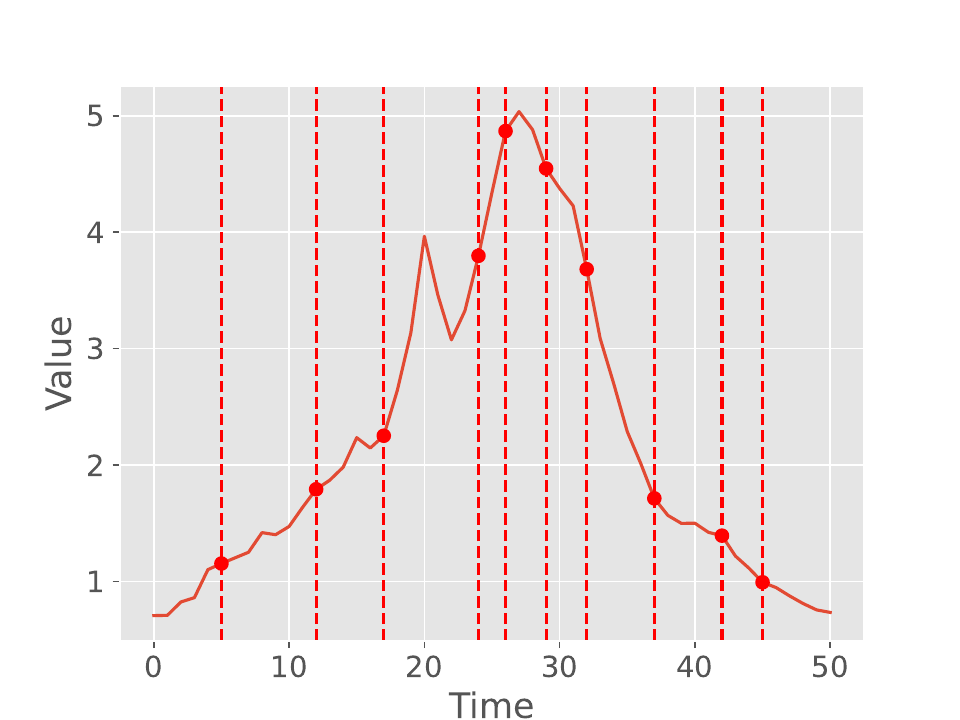}
		\caption{Flu-US}
		\label{fig:sub1}
	\end{subfigure}
	\hfill
	\begin{subfigure}[b]{0.32\textwidth}
		\centering
		\includegraphics[width=\textwidth]{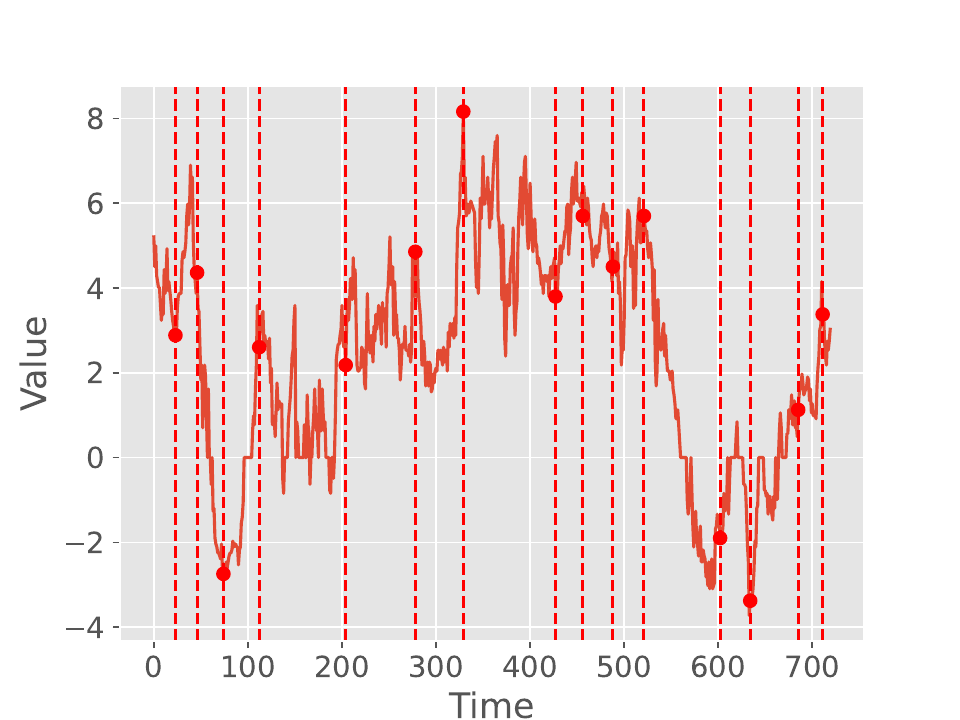}
		\caption{ETT1}
		\label{fig:sub2}
	\end{subfigure}
	\hfill
	\begin{subfigure}[b]{0.32\textwidth}
		\centering
		\includegraphics[width=\textwidth]{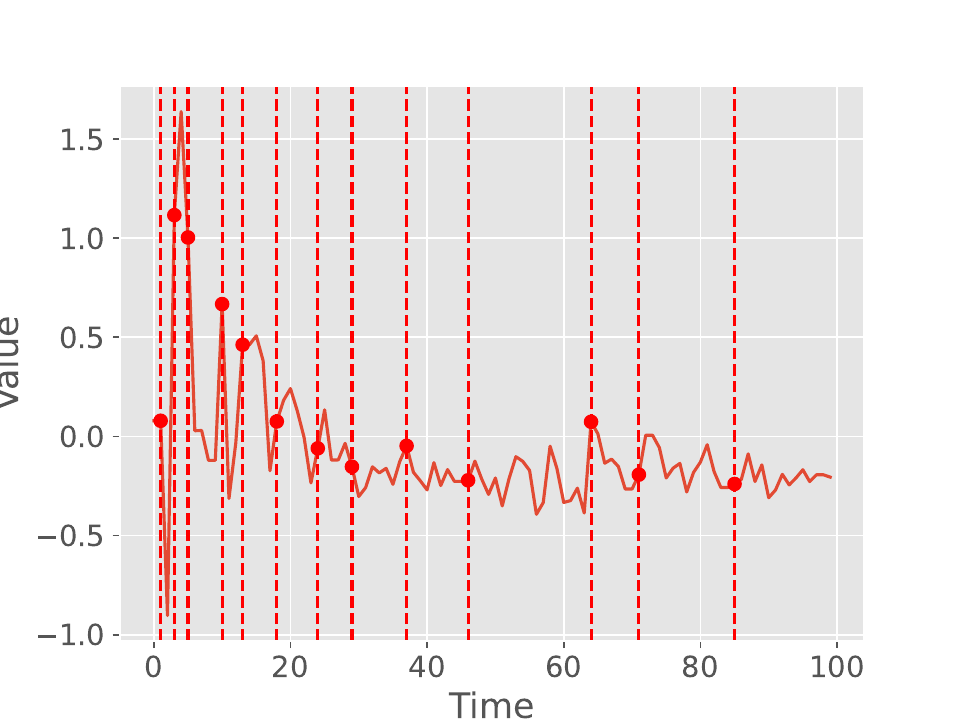}
		\caption{BasicMotions}
		\label{fig:sub3}
	\end{subfigure}
	\caption{Segmentation learned by \model}
	\label{fig:segment}
\end{figure}

\paragraph{Segments generated by \model}
We also visualized the segments in Fig. \ref{fig:segment}. We observe that the segment sizes are smaller in regions of high variance or important parts of the time-series such as peak of the epidemic whereas simpler trends have longer segments which matches our intuition.

\section{Conclusion}
We make a significant contribution towards general pre-trained models for time-series analysis tasks replicating the success of large pre-trained models in language and vision domains.
We introduce \model, a general pre-trained model that provides state-of-art performance on a wide range of forecasting and classification tasks from varied domains and applications. \model provides similar to or better performance to state-of-art domain-specific models in applications such as epidemiology, energy, traffic, and economics and significantly beats recent time-series foundational models.
We also observe that \model required significantly lesser pre-training and training data to reach optimal performance compared to other baselines in most benchmarks.

Our work mainly focuses on the important challenge of providing semantically meaningful inputs to the model that caters to learning time-series segmentation strategies specific to each domain.
This is crucial when pre-training on diverse datasets, a key challenge for time-series data.
The underlying model architecture is a straightforward transformer encoder that uses
well-known masking techniques for self-supervised pre-training.
Therefore, our method can be extended to leverage novel time-series model
architectures and SSL methods.
Extending our methods to provide calibrated forecasts that provide reliable uncertainty measures is also another important direction of research.
We can also extend it to leverage multimodal datasets like text that provide important contextual information about the dataset~\cite{liu2024timemmd}.

Since our model can be applied to any generic time-series analysis tasks including those in critical domains such as public health, medicine, economics, etc., important steps need to be taken to address potential misuse of the
our methods such as testing for fairness, data quality issues, ethical implications of predictions, etc.
\section*{Acknowledgements}
This paper was supported in part bythe NSF (Expeditions CCF-1918770, CAREER IIS-2028586, Medium IIS-1955883, Medium IIS-2403240, Medium IIS-2106961, PIPP CCF-2200269), CDC MInD program, Meta faculty gifts, and funds/computing resources from Georgia Tech.

\bibliographystyle{ACM-Reference-Format}
\bibliography{references}
\newpage
\appendix

{\Large \bf Appendix for Large Pre-trained time series models for cross-domain
	Time series analysis tasks}
\section{Hyperparameters}

The model is run on Intel Xeon CPU with 64 cores and 128 GB RAM. We use a single A100 GPU with 80GB memory.
For GRU we use a single hidden layer of 50 hidden units. Dimension of $\mathbf{v}$ is also 50.
The transformer architecture consists of 10 layers with 8 attention heads each. For forecasting tasks, we train a separate decoder module with 4 more layers during fine-tuning
whereas for classification we aggregate the embeddings $\{e_i\}_{i=1}^R$ of the last transformer layer and feed them into a single linear layer that provides logits for all classes.
The SSL pre-training was done till convergence via early stopping with patience of 1000 epochs. We observed that \model takes 5000-8000 epochs to finish pre-training which takes
around 3-4 hours. (Note that pre-training is a one-time step and downstream fine-tuning takes much less time and epochs).
For both pre-training and fine-tuning, we used the Adam optimizer with a learning rate of 0.001.
The hyperparameters are tuned sparingly for both \model and baselines from their default settings.
For \randmask, we found the optimal $\gamma = 0.4$, and for \lastmask $\gamma=0.2$ was optimal.

\section{Data efficiency}
\begin{table*}[h]
	\caption{Average training time (minutes) till convergence for \model and neural baselines. \model-TB shows the time taken by \model to reach performance of top baseline (in benchmarks where \model outperforms it). Since some baselines are specific to forecasting or classification and we do not beat the state-of-art in few benchmarks we designate these cells in the table as NA.}

	\label{tab:traintime}
	\scalebox{0.85}{
		\begin{tabular}{c|cccccccc}
			Model      & Flu-US        & ETT2          & PEM-Bays      & NY-Bike       & Nasdaq        & M3            & BasicMotions & EigenWorms   \\ \hline
			Informer   & 27.3          & 25.5          & 45.1          & 49.7          & 27.1          & 49.6          & 17.5         & 14.3         \\
			Autoformer & 19.5          & 29.3          & 49.5          & 55.2          & 18.5          & 45.1          & 11.9         & 19.7         \\
			N-HITS     & 15.4          & 22.5          & 36.1          & 49.3          & 26.4          & 47.5          & NA           & NA           \\
			PatchTST   & 12.9          & 29.5          & 36.4          & 45.7          & 18.2          & 49.4          & NA           & NAg          \\
			MICN       & 17.6          & 15.7          & 39.7          & 41.1          & 19.2          & 33.9          & NA           & NA           \\
			TimesNet   & 15.4          & 19.7          & 37.4          & 46.3          & 24.1          & 36.5          & 9.4          & 11.5         \\
			STEP       & 25.4          & 34.1          & 52.7          & 74.3          & 29.7          & 52.8          & NA           & NA           \\
			EpiFNP     & 22.5          & 39.5          & 41.1          & 39.1          & 21.6          & 97.6          & NA           & NA           \\
			ColaGNN    & 34.7          & 33.6          & 53.1          & 47.6          & 32.1          & 72.2          & NA           & NA           \\
			TARNet     & NA            & NA            & NA            & NA            & NA            & NA            & 13.7         & 9.4          \\
			TS2Vec     & 29.3          & 28.2          & 41.9          & 41.9          & 29.8          & 67.4          & 9.3          & 13.2         \\
			TS-TCC     & 21.7          & 23.7          & 46.3          & 44.3          & 25.3          & 55.8          & 12.7         & 11.1         \\ \hline
			\model     & \textbf{12.2} & 19.3          & 41.9          & \textbf{37.5} & \textbf{17.3} & \textbf{31.2} & \textbf{6.1} & 12.7         \\
			\model-TB  & NA            & \textbf{12.5} & \textbf{29.6} & \textbf{32.9} & NA            & \textbf{23.7} & \textbf{6.1} & \textbf{8.1}
		\end{tabular}
	}
\end{table*}

\begin{figure*}[h]
	\centering
	\begin{subfigure}{0.24\textwidth}
		\includegraphics[width=\linewidth]{EffImgs/Flu-US.pdf}
	\end{subfigure}
	\hfill
	\begin{subfigure}{0.24\textwidth}
		\includegraphics[width=\linewidth]{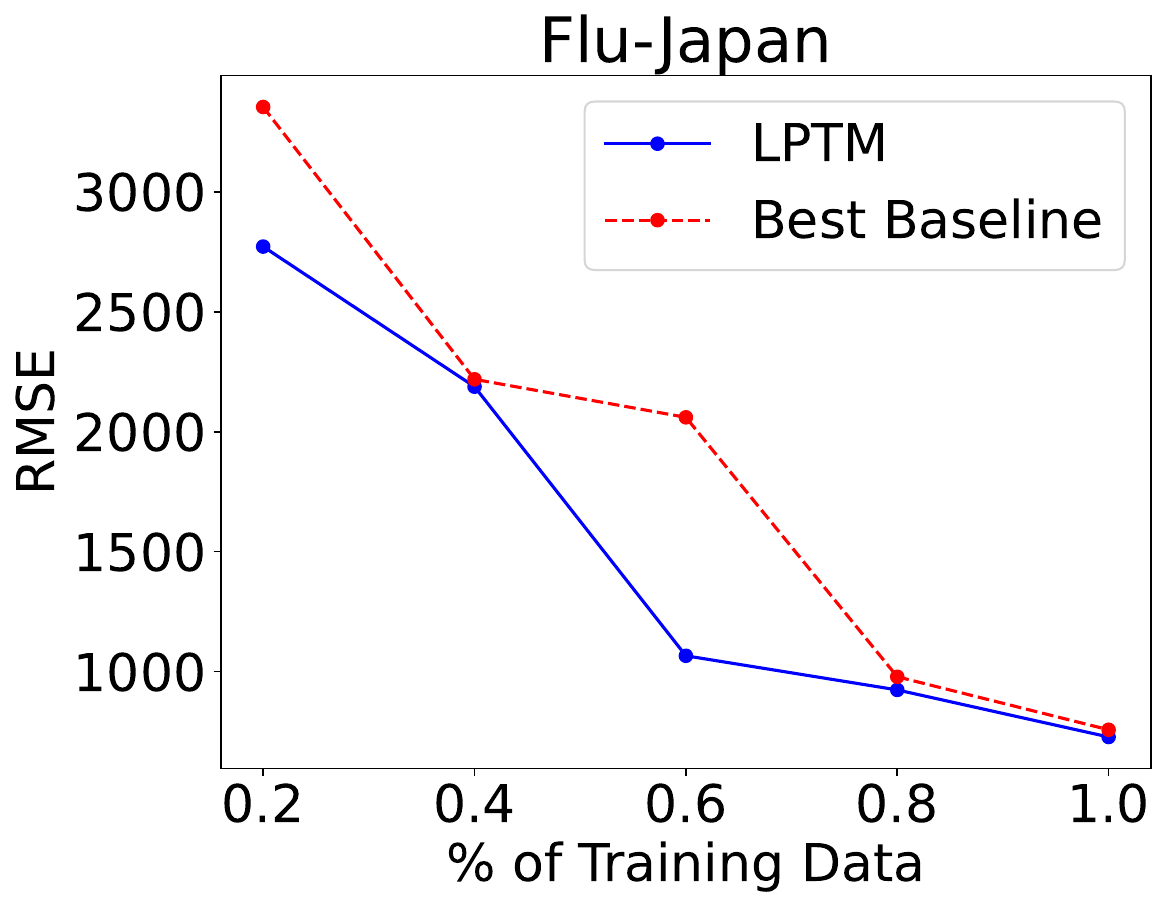}
	\end{subfigure}
	\hfill
	\begin{subfigure}{0.24\textwidth}
		\includegraphics[width=\linewidth]{EffImgs/ETT2.pdf}
	\end{subfigure}
	\hfill
	\begin{subfigure}{0.24\textwidth}
		\includegraphics[width=\linewidth]{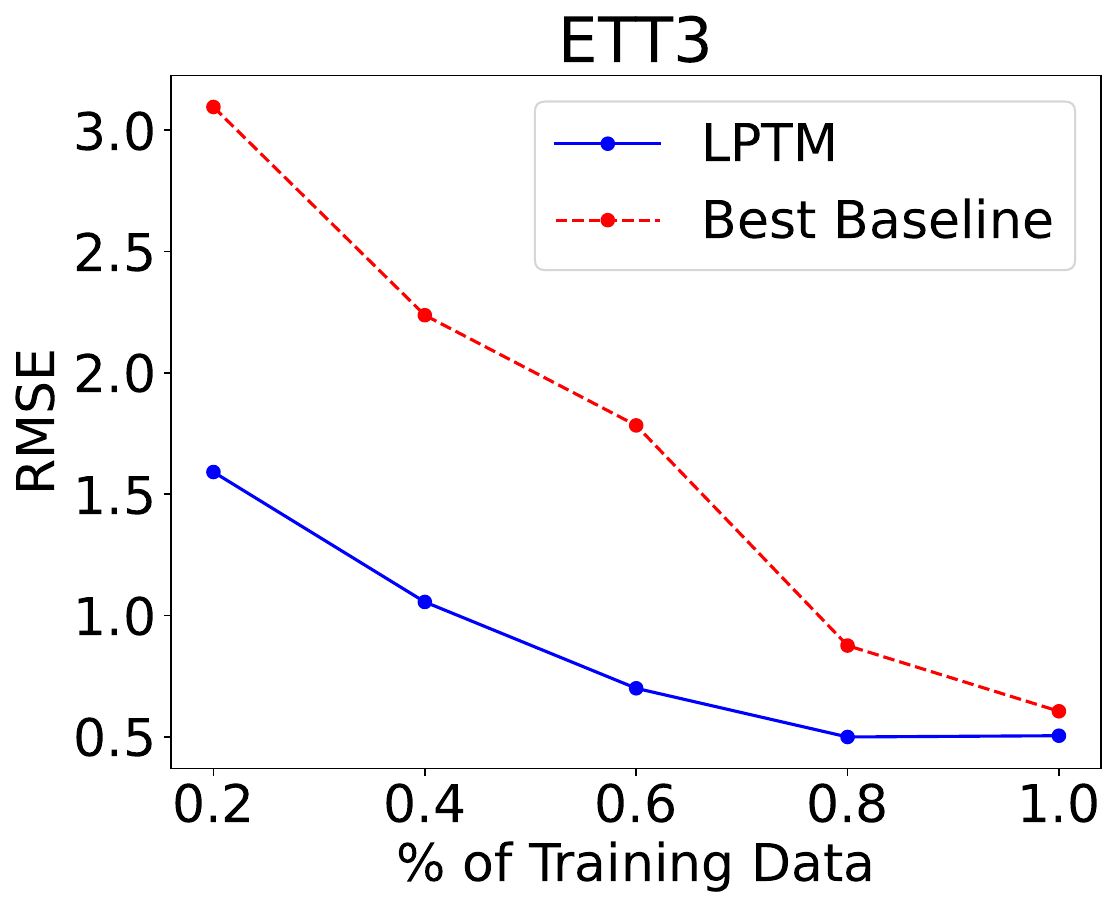}
	\end{subfigure}

	\begin{subfigure}{0.24\textwidth}
		\includegraphics[width=\linewidth]{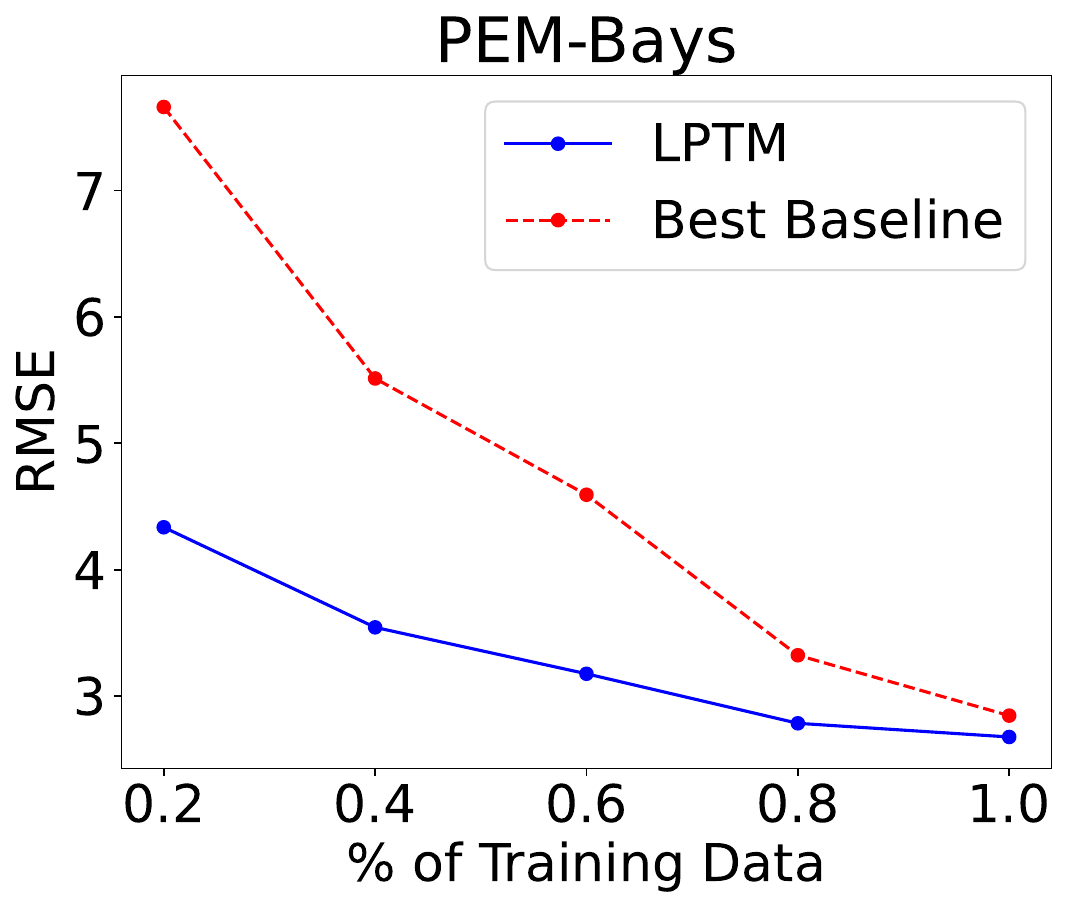}
	\end{subfigure}
	\hfill
	\begin{subfigure}{0.24\textwidth}
		\includegraphics[width=\linewidth]{EffImgs/NY-Bike.pdf}
	\end{subfigure}
	\begin{subfigure}{0.24\textwidth}
		\includegraphics[width=\linewidth]{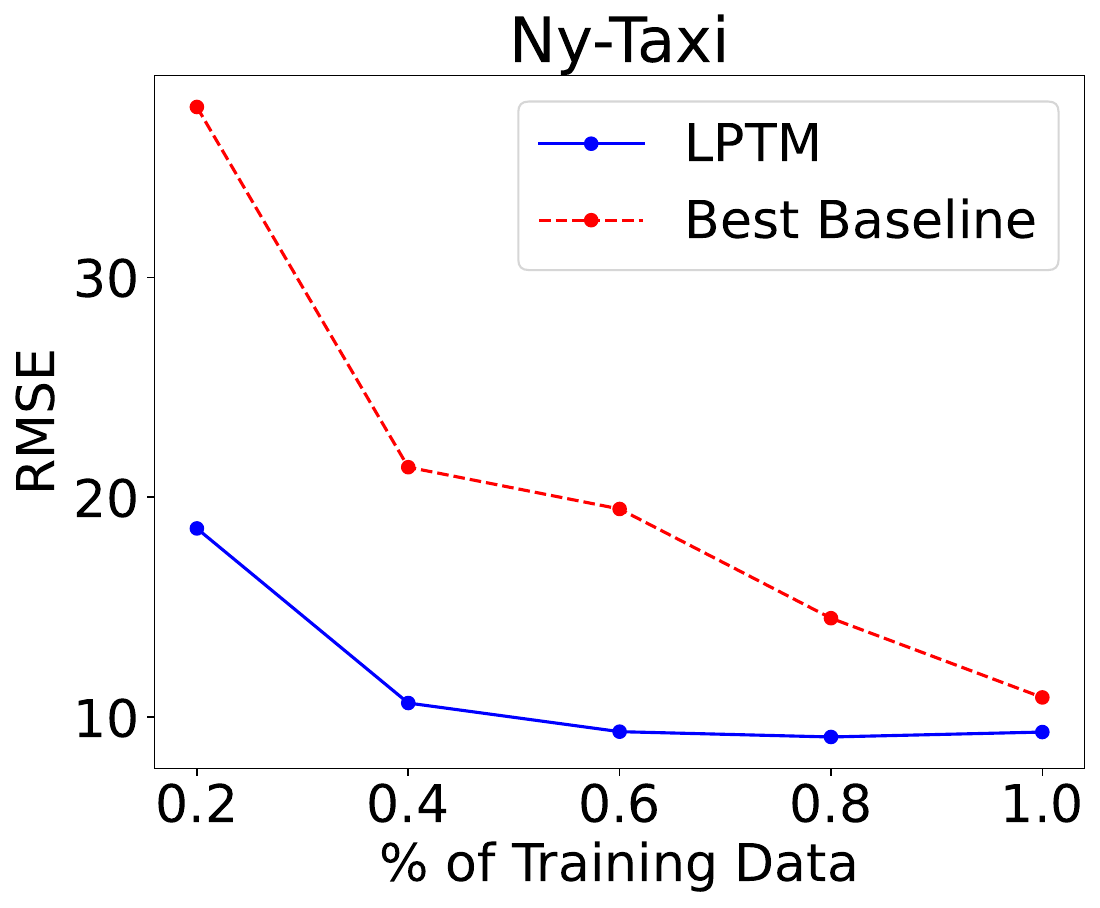}
	\end{subfigure}
	\hfill
	\begin{subfigure}{0.24\textwidth}
		\includegraphics[width=\linewidth]{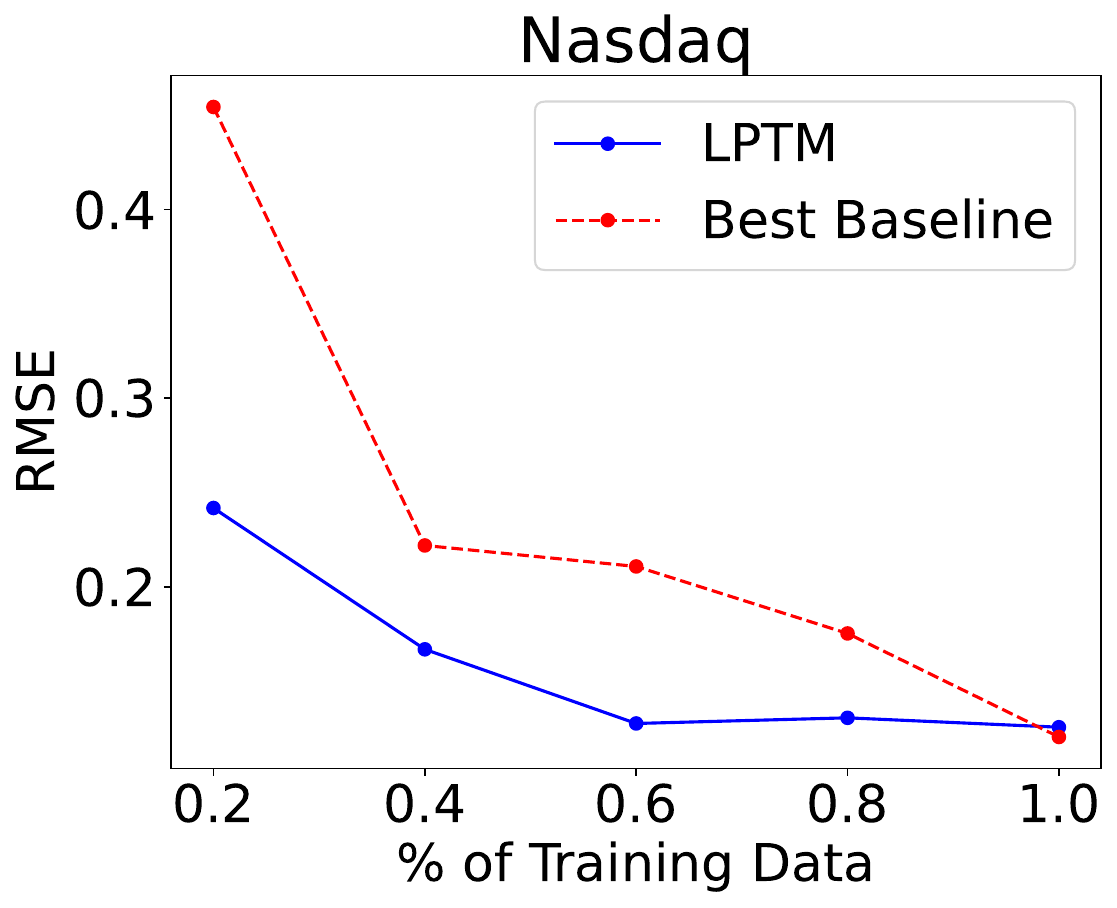}
	\end{subfigure}


	\begin{subfigure}{0.24\textwidth}
		\includegraphics[width=\linewidth]{EffImgs/M3.pdf}
	\end{subfigure}
	\hfill
	\begin{subfigure}{0.24\textwidth}
		\includegraphics[width=\linewidth]{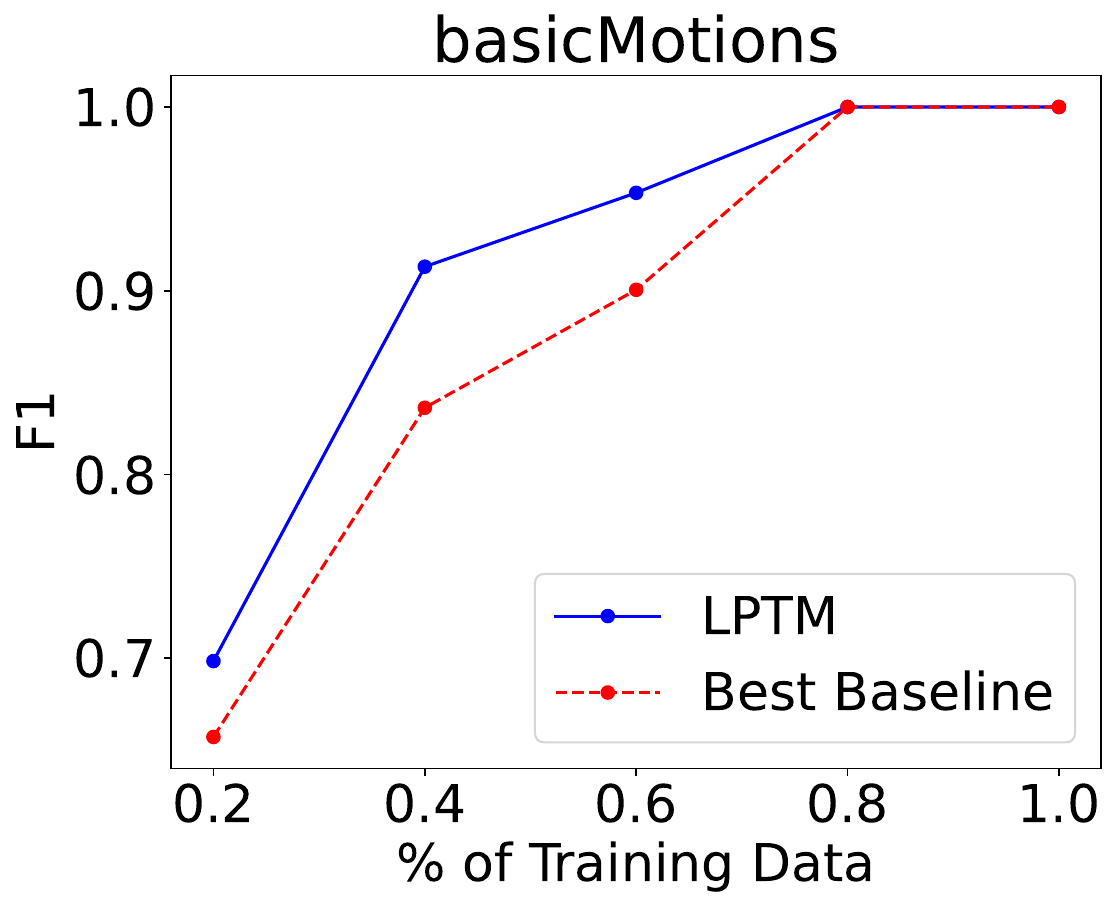}
	\end{subfigure}
	\hfill
	\begin{subfigure}{0.24\textwidth}
		\includegraphics[width=\linewidth]{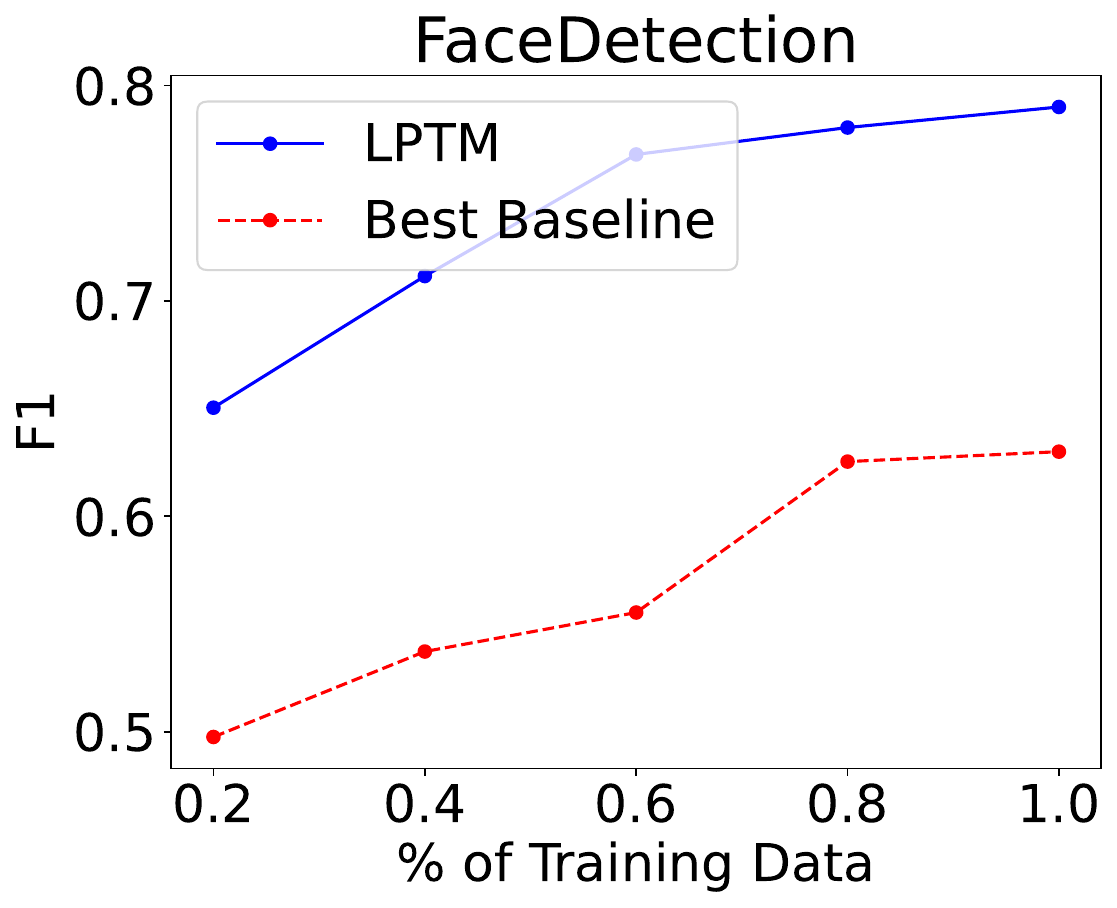}
	\end{subfigure}
	\hfill
	\begin{subfigure}{0.24\textwidth}
		\includegraphics[width=\linewidth]{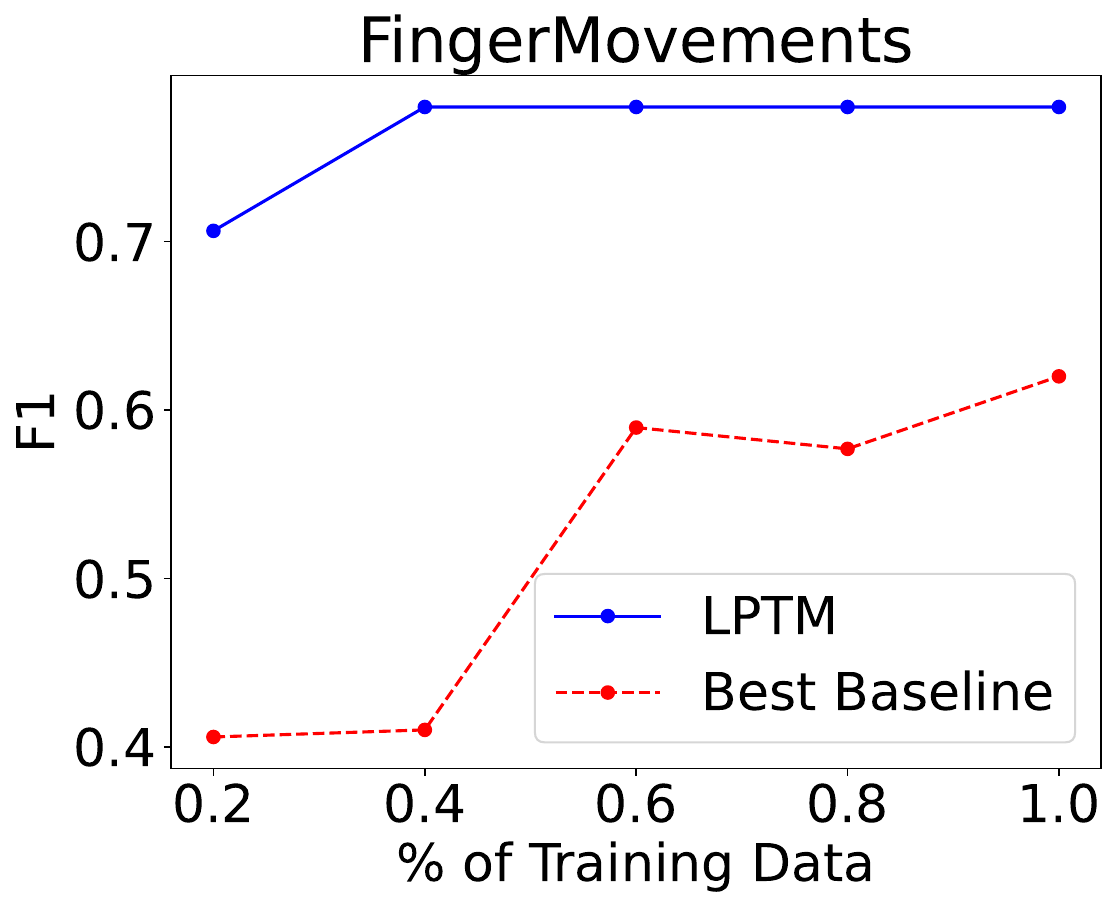}
	\end{subfigure}
	\begin{subfigure}{0.24\textwidth}
		\includegraphics[width=\linewidth]{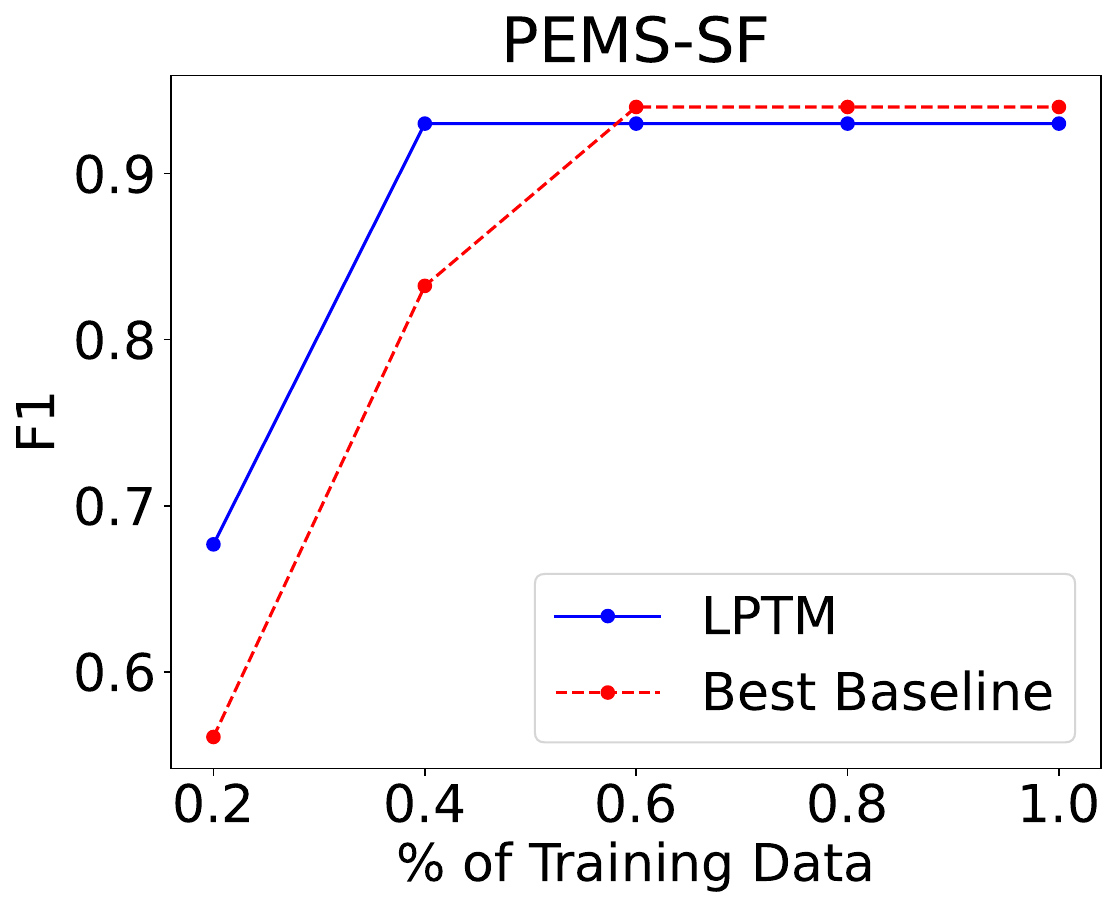}
	\end{subfigure}
	\hfill
	\begin{subfigure}{0.24\textwidth}
		\includegraphics[width=\linewidth]{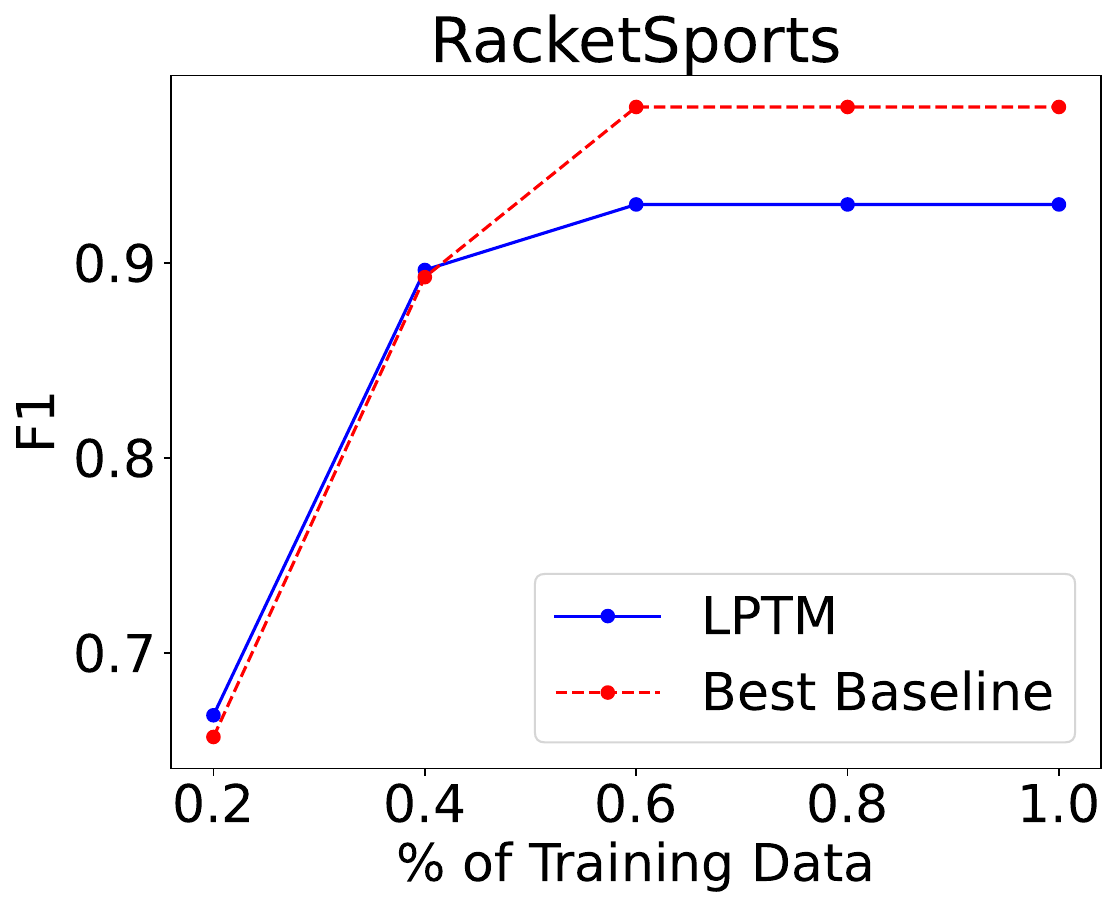}
	\end{subfigure}
	\hfill
	\begin{subfigure}{0.24\textwidth}
		\includegraphics[width=\linewidth]{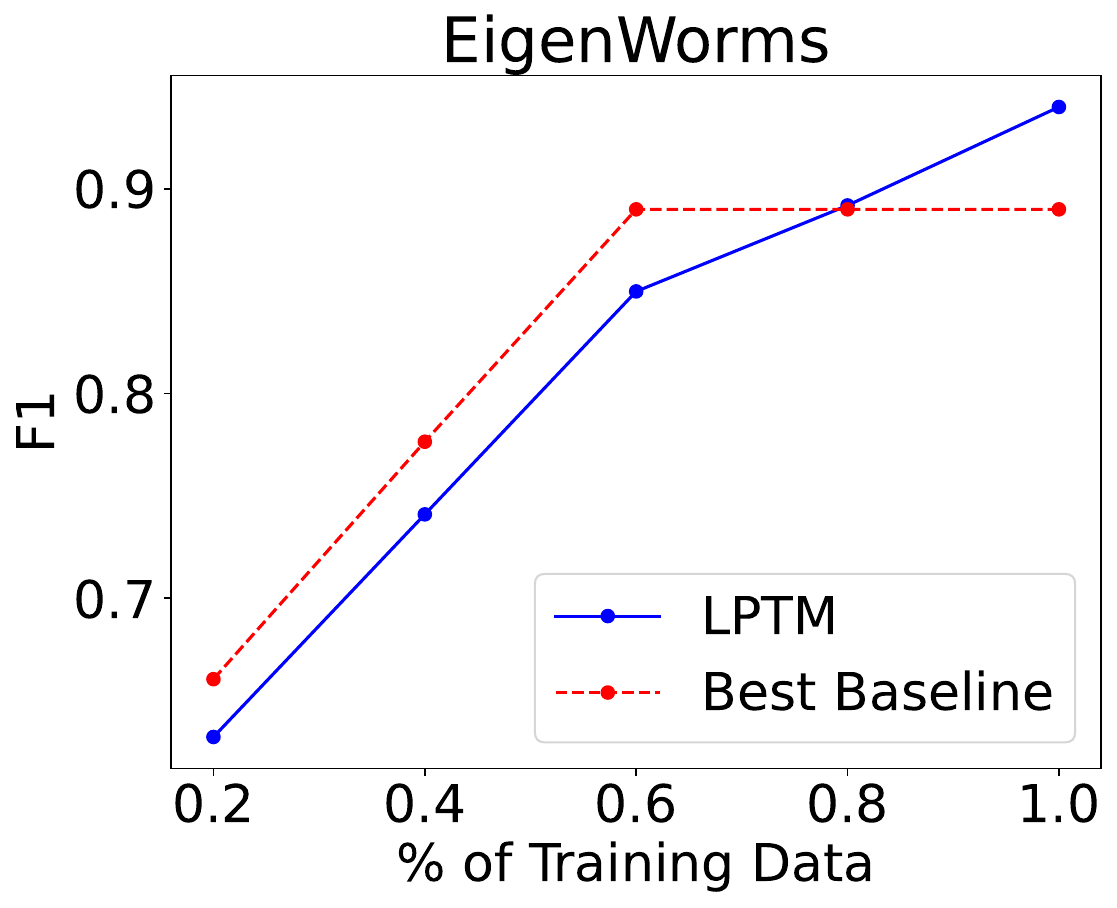}
	\end{subfigure}

	\caption{Performance of \model and best baseline with varying fractions of training data.  In most cases \model significantly outperforms baselines with lower amount of data.}
\end{figure*}
\clearpage


\section{Effect of SSL hyperparameter $\gamma$}
\begin{figure*}[h]
	\centering
	\includegraphics[width=.9\textwidth]{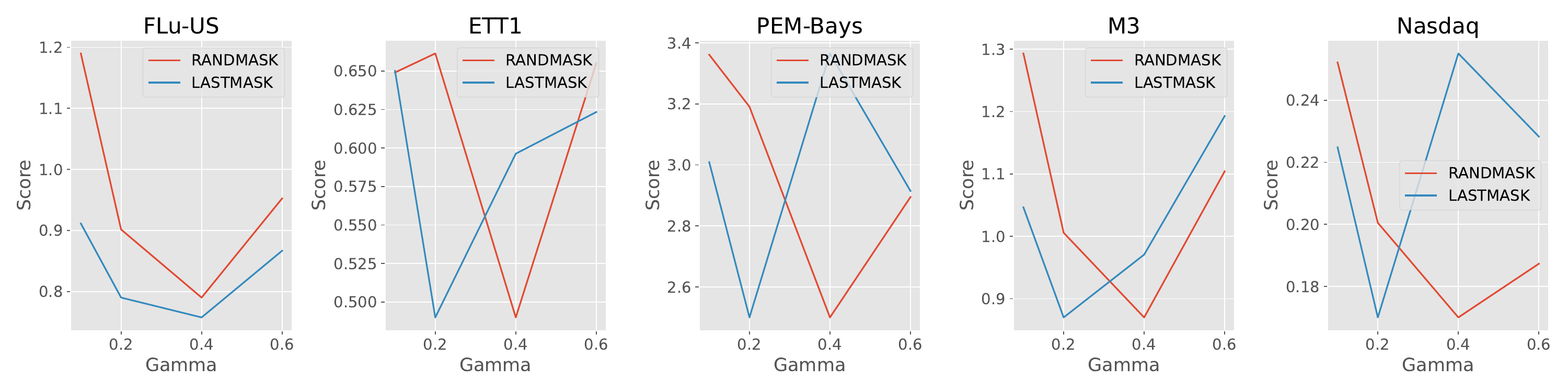}
	\caption{Effect of $\gamma$ on performance(RMSE) for different benchmarks}
	\label{fig:gamma}
\end{figure*}

\clearpage

\begin{table}[h]
	\caption{Std. dev across 10 runs}
	\centering
	\scalebox{0.8}{
		\begin{tabular}{l|rrrrrrrrr}
			\toprule
			Model             & Flu-US   & Flu-Japan & ETT1     & ETT2     & PEM-Bays & NY-Bike  & NY-Taxi  & Nasdaq   & M4       \\
			\midrule
			AutoARIMA         & 0.043180 & 23.434077 & 0.014220 & 0.035201 & 0.056066 & 0.044572 & 0.151146 & 0.025181 & 0.024060 \\
			Informer          & 0.044072 & 23.918021 & 0.014514 & 0.035927 & 0.057224 & 0.045492 & 0.154268 & 0.025701 & 0.024557 \\
			Autoformer        & 0.042310 & 22.961906 & 0.013934 & 0.034491 & 0.054936 & 0.043674 & 0.148101 & 0.024673 & 0.023575 \\
			PatchTST          & 0.043320 & 23.509948 & 0.014266 & 0.035314 & 0.056248 & 0.044716 & 0.151636 & 0.025262 & 0.024138 \\
			N-HITS            & 0.048238 & 26.178970 & 0.015886 & 0.039324 & 0.062633 & 0.049793 & 0.168850 & 0.028130 & 0.026878 \\
			MICN              & 0.030538 & 16.573161 & 0.010057 & 0.024895 & 0.039651 & 0.031522 & 0.106894 & 0.017808 & 0.017016 \\
			TimesNet          & 0.021264 & 11.539913 & 0.007003 & 0.017334 & 0.027609 & 0.021949 & 0.074431 & 0.012400 & 0.011848 \\
			LLM-Time          & 0.040279 & 21.859627 & 0.013265 & 0.032836 & 0.052299 & 0.041577 & 0.140991 & 0.023489 & 0.022443 \\
			TimesFM           & 0.034938 & 18.961321 & 0.011506 & 0.028482 & 0.045365 & 0.036065 & 0.122298 & 0.020375 & 0.019468 \\
			Lag-LLAMA         & 0.028790 & 15.624783 & 0.009481 & 0.023470 & 0.037382 & 0.029719 & 0.100777 & 0.016789 & 0.016042 \\
			Chronos           & 0.040037 & 21.728648 & 0.013185 & 0.032639 & 0.051986 & 0.041328 & 0.140146 & 0.023348 & 0.022309 \\
			STEP              & 0.028283 & 15.349319 & 0.009314 & 0.023056 & 0.036723 & 0.029195 & 0.099001 & 0.016493 & 0.015759 \\
			EpiFNP            & 0.026098 & 14.163743 & 0.008595 & 0.021275 & 0.033887 & 0.026940 & 0.091354 & 0.015219 & 0.014542 \\
			ColaGNN           & 0.038447 & 20.865267 & 0.012661 & 0.031342 & 0.049920 & 0.039686 & 0.134578 & 0.022421 & 0.021422 \\
			TS2Vec            & 0.032028 & 17.381828 & 0.010548 & 0.026109 & 0.041586 & 0.033061 & 0.112110 & 0.018677 & 0.017846 \\
			SimMTM            & 0.053732 & 29.160611 & 0.017695 & 0.043802 & 0.069767 & 0.055464 & 0.188081 & 0.031334 & 0.029939 \\
			TS-TCC            & 0.044399 & 24.095686 & 0.014622 & 0.036194 & 0.057649 & 0.045830 & 0.155414 & 0.025892 & 0.024739 \\
			LPTM              & 0.048677 & 26.417645 & 0.016031 & 0.039682 & 0.063204 & 0.050247 & 0.170390 & 0.028387 & 0.027123 \\
			LPTM-NoSegment    & 0.045970 & 24.948198 & 0.015139 & 0.037475 & 0.059689 & 0.047452 & 0.160912 & 0.026808 & 0.025614 \\
			LPTM-NoPreTrain   & 0.058905 & 31.968143 & 0.019399 & 0.048020 & 0.076484 & 0.060804 & 0.206190 & 0.034351 & 0.032822 \\
			LPTM-NoLinProb    & 0.025968 & 14.092985 & 0.008552 & 0.021169 & 0.033718 & 0.026805 & 0.090898 & 0.015143 & 0.014469 \\
			LPTM-OnlyRandMask & 0.039517 & 21.446157 & 0.013014 & 0.032214 & 0.051310 & 0.040791 & 0.138324 & 0.023045 & 0.022019 \\
			LPTM-OnlyLastMask & 0.058067 & 31.513665 & 0.019123 & 0.047337 & 0.075397 & 0.059940 & 0.203258 & 0.033863 & 0.032355 \\
			\bottomrule
		\end{tabular}
		\vspace{-0.2in}
	}
\end{table}

\begin{table}[h]
	\centering
	\scalebox{0.8}{
		\begin{tabular}{|l|r|}
			\hline
			Model             & Score     \\
			\hline
			AutoARIMA         & 21.388889 \\
			Informer          & 14.055556 \\
			Autoformer        & 12.000000 \\
			PatchTST          & 6.833333  \\
			N-HITS            & 11.388889 \\
			MICN              & 7.277778  \\
			TimesNet          & 11.111111 \\
			LLM-Time          & 12.777778 \\
			TimesFM           & 12.277778 \\
			Lag-LLAMA         & 17.444444 \\
			Chronos           & 13.888889 \\
			STEP              & 9.111111  \\
			EpiFNP            & 14.666667 \\
			ColaGNN           & 16.500000 \\
			TS2Vec            & 19.111111 \\
			SimMTM            & 15.611111 \\
			TS-TCC            & 18.111111 \\
			LPTM              & 2.500000  \\
			LPTM-NoSegment    & 12.055556 \\
			LPTM-NoPreTrain   & 10.444444 \\
			LPTM-NoLinProb    & 6.222222  \\
			LPTM-OnlyRandMask & 7.555556  \\
			LPTM-OnlyLastMask & 3.666667  \\
			\hline
		\end{tabular}}
	\caption{Mean rank of models in Table 2}
	\label{tab:my_label}
\end{table}

\newpage

\newpage
\section*{NeurIPS Paper Checklist}


\begin{enumerate}

	\item {\bf Claims}
	\item[] Question: Do the main claims made in the abstract and introduction accurately reflect the paper's contributions and scope?
	\item[] Answer: \answerYes{} 
	\item[] Justification: The contributions listed in the Introduction are clearly expanded in Methodology and Results

	\item {\bf Limitations}
	\item[] Question: Does the paper discuss the limitations of the work performed by the authors?
	\item[] Answer: \answerYes{} 
	\item[] Justification: Conclusions discuss limitiations such as not applicable to probabilistic and multivariate forecasts and potential extensions as future work.

	\item {\bf Theory Assumptions and Proofs}
	\item[] Question: For each theoretical result, does the paper provide the full set of assumptions and a complete (and correct) proof?
	\item[] Answer: \answerNA{} 
	\item[] Justification: No theoretical results.

	\item {\bf Experimental Result Reproducibility}
	\item[] Question: Does the paper fully disclose all the information needed to reproduce the main experimental results of the paper to the extent that it affects the main claims and/or conclusions of the paper (regardless of whether the code and data are provided or not)?
	\item[] Answer: \answerYes{} 
	\item[] Justification: We provide model architecture and training details in methodology, link to the anonymized repo and hyperparameters in Appendix A.

	\item {\bf Open access to data and code}
	\item[] Question: Does the paper provide open access to the data and code, with sufficient instructions to faithfully reproduce the main experimental results, as described in supplemental material?
	\item[] Answer: \answerYes{} 
	\item[] Justification: Data and code available in \url{https://github.com/AdityaLab/LPTM/}

	\item {\bf Experimental Setting/Details}
	\item[] Question: Does the paper specify all the training and test details (e.g., data splits, hyperparameters, how they were chosen, type of optimizer, etc.) necessary to understand the results?
	\item[] Answer: \answerYes{} 
	\item[] Justification: Specific details on hyperparameters are in Appendix A. Training details are in Section 3.3. Experiment setup, dataset splits are in Sec 5.

	\item {\bf Experiment Statistical Significance}
	\item[] Question: Does the paper report error bars suitably and correctly defined or other appropriate information about the statistical significance of the experiments?
	\item[] Answer: \answerYes{} 
	\item[] Justification: We perform multiple (10) runs of each experiment and report the average scores. We highlight the instances where \model in statistically significantly better than all baselines.

	\item {\bf Experiments Compute Resources}
	\item[] Question: For each experiment, does the paper provide sufficient information on the computer resources (type of compute workers, memory, time of execution) needed to reproduce the experiments?
	\item[] Answer: \answerYes{} 
	\item[] Justification: We mention the compute resources used in Appendix A and training time in Appendix B.

	\item {\bf Code Of Ethics}
	\item[] Question: Does the research conducted in the paper conform, in every respect, with the NeurIPS Code of Ethics \url{https://neurips.cc/public/EthicsGuidelines}?
	\item[] Answer: \answerYes{} 
	\item[] Justification: We went through and confirm the paper follows all the statements in the code of ethics.

	\item {\bf Broader Impacts}
	\item[] Question: Does the paper discuss both potential positive societal impacts and negative societal impacts of the work performed?
	\item[] Answer: \answerNo{} 
	\item[] Justification: Our work is a general purpose time-series model that can be utilized towards wide range of applications. Care should be taken to assess the ethics of the decision taken by the predictions, clean the data for any sensitive information and ensure fairness and positive utility in the resulting outcome.

	\item {\bf Safeguards}
	\item[] Question: Does the paper describe safeguards that have been put in place for responsible release of data or models that have a high risk for misuse (e.g., pretrained language models, image generators, or scraped datasets)?
	\item[] Answer: \answerNo{} 
	\item[] Justification: Since this work is a general-purpose foundational work on time-series and we do not focus on any potentially sensitive applications, providing safeguards for release of our work is not applicable.

	\item {\bf Licenses for existing assets}
	\item[] Question: Are the creators or original owners of assets (e.g., code, data, models), used in the paper, properly credited and are the license and terms of use explicitly mentioned and properly respected?
	\item[] Answer: \answerYes{} 
	\item[] Justification: All assets in code are either created by authors or are derived from open-source works and are cited appropriately.

	\item {\bf New Assets}
	\item[] Question: Are new assets introduced in the paper well documented and is the documentation provided alongside the assets?
	\item[] Answer: \answerYes{} 
	\item[] Justification: The code is the only original asset which is released as an anonymous repo linked in the paper. We will make a public release of this on acceptance.

	\item {\bf Crowdsourcing and Research with Human Subjects}
	\item[] Question: For crowdsourcing experiments and research with human subjects, does the paper include the full text of instructions given to participants and screenshots, if applicable, as well as details about compensation (if any)?
	\item[] Answer: \answerNA{} 
	\item[] Justification: No crowdsourcing research.

	\item {\bf Institutional Review Board (IRB) Approvals or Equivalent for Research with Human Subjects}
	\item[] Question: Does the paper describe potential risks incurred by study participants, whether such risks were disclosed to the subjects, and whether Institutional Review Board (IRB) approvals (or an equivalent approval/review based on the requirements of your country or institution) were obtained?
	\item[] Answer: \answerNA{} 
	\item[] Justification: This work doesn't deal with research requiring IRB review.

\end{enumerate}

\end{document}